\documentclass[10pt,twocolumn,letterpaper]{article}

\usepackage[pagenumbers]{cvpr}

\usepackage{wrapfig}
\usepackage{graphicx}
\usepackage{booktabs}
\usepackage{multirow}
\usepackage{tabularray}
\usepackage[innercaption]{sidecap}
\usepackage{colortbl}
\usepackage[accsupp]{axessibility}
\usepackage{float}
\usepackage{makecell}
\usepackage{caption}
\usepackage{stfloats}
\IfFileExists{soul.sty}{\usepackage{soul}}{}
\usepackage{xcolor}
\usepackage{xspace}
\usepackage{arydshln}
\usepackage{tabularx}
\IfFileExists{soul.sty}{\sethlcolor{cyan!15}}{}

\newcommand{\ours}{\cellcolor{MidnightBlue!8}}

\newcommand{\tablestyle}[1]{%
  \small
  \setlength{\aboverulesep}{2pt}%
  \setlength{\belowrulesep}{2pt}%
  \renewcommand{\arraystretch}{1}%
  \setlength{\tabcolsep}{#1}%
  \vspace{-2mm}%
}

\newcommand{\method}{FoldSketch\xspace}
\definecolor{cvprblue}{rgb}{0.21,0.49,0.74}
\usepackage[pagebackref,breaklinks,colorlinks,allcolors=cvprblue]{hyperref}

\def\confName{CVPR}
\def\confYear{2027}

\title{\vspace{-2mm}FoldSketch: Drawings That Unfold When Folded\vspace{-4mm}}

\author{
\textbf{\href{https://chaitron.github.io}{Chaitat Utintu}}$^{1,2}$, \hspace{.3cm}
\textbf{\href{https://thecoder1012.github.io}{Arkaprabha Basu}}$^{1}$, \hspace{.3cm}
\textbf{\href{https://www.surrey.ac.uk/people/anjan-dutta}{Anjan Dutta}}$^{3}$, \hspace{.3cm}
\textbf{\href{https://www.surrey.ac.uk/people/yi-zhe-song}{Yi-Zhe Song}}$^{1}$\\
$^{1}$ SketchX, Surrey Institute for People-Centred Artificial Intelligence (PAI), University of Surrey\\
$^{2}$ UKRI AI: CDT in AI for Digital Media Inclusion \hspace{.4cm} $^{3}$ CVSSP, University of Surrey\\
{\texttt{\{chaitat.u, a.basu, anjan.dutta, y.song\}@surrey.ac.uk}}
}

\begin{document}

\twocolumn[{%
\renewcommand\twocolumn[1][]{#1}%
\maketitle
    \captionsetup{type=figure}
    \vspace{-0.6cm}
    \centering
    \includegraphics[width=\textwidth]{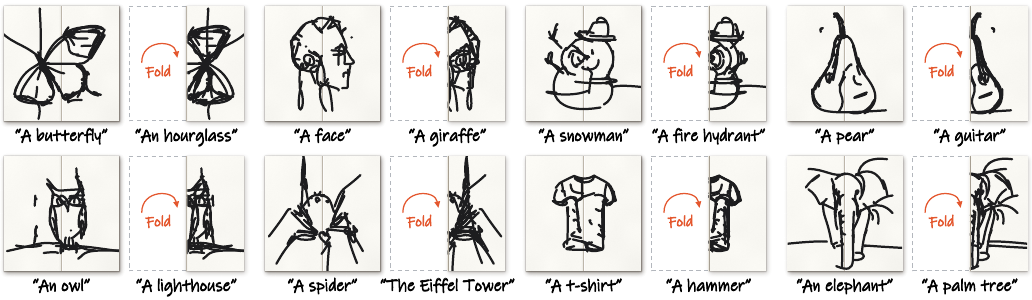} \\[-0.4cm]
    \caption{\textbf{One sketch, two readings.} Each pair shows one sheet of vector strokes, flat and then folded. Flat, it reads as the first prompt. Folding its left half over the right one, as on tracing paper, turns it into the second. The dashed outline marks where the folded half lay.}
    \label{fig:teaser}
    \vspace{1mm}
}]

\begin{abstract}
Fold the owl in half and it becomes a lighthouse. For half a century the back page of MAD magazine ended this way, with a fold-in that Al Jaffee planned line by line so that the open page and the folded one each made sense. We ask whether a machine can draw one, and in a stricter form than his: Jaffee's fold hides the middle of the page, ours hides nothing, so every stroke stays in view and must earn its place in both pictures. A fold is a reflection about the crease, so the folded drawing is the flat drawing superimposed on its own mirror image. It piles ink up instead of rearranging it: the map is many-to-one, and the inverse-view denoising behind visual anagrams no longer applies. Sparse strokes are what make such illusions legible: the strokes of two layers pass between each other where two folded photographs mix at every pixel. \method takes a prompt per reading and a set of creases and returns one SVG with the folds that reveal each reading. A composable fold operator maps a set of strokes to strokes, so any sequence of creases is one map applied in turn. Shared-stroke distillation lets every stroke be shaped by each reading it takes part in, reading balance shifts weight towards the weaker reading, and a crease loss keeps the two layers of a fold from drawing over each other. FoldScore rates a sheet by its worst reading, so that a sheet counts only if every reading works. On 50 pairs of concepts, \method scores higher than every vector sketch method we compare with and level with a raster illusion method that inks 54\% of the sheet where \method inks 6\%.
\end{abstract}

\section{Introduction}
\label{sec:intro}
For fifty‑six years the back page of MAD magazine carried a picture that became another picture when you folded it. Al Jaffee drew nearly all of those fold‑ins~\cite{jaffee2011madfoldin}, planning each line to serve the open page and the folded one at once. The fold-in belongs to a family of pictures that hold more than one drawing: the duck-rabbit~\cite{jastrow1899mindseye} switches without a line changing, hybrid images~\cite{oliva2006hybrid} change with viewing distance, shadow art~\cite{mitra2009shadowart} and wire art~\cite{hsiao2018wireart,qu2024dreamwire} show a different picture from each direction. What sets the fold-in apart is that the viewer makes the second picture with their hands. The illusion lives in the paper, not on a screen. We ask whether such drawings can be generated rather than drawn, and in a stricter form than Jaffee's: can one vector sketch be built so that each fold of the sheet reveals a drawing of its own (\cref{fig:teaser})?

Diffusion models now generate multi-view illusions~\cite{burgert2024diffusionillusions,geng2024visualanagrams,geng2024factorized,chang2025lookingglass}. Visual Anagrams~\cite{geng2024visualanagrams} denoises several views of one noisy image, each with its own prompt, maps the noise estimates back through the inverse views and averages them. This is valid for orthogonal views, such as rotations, flips and pixel permutations, which rearrange the image without loss. A fold has no inverse. It does not rearrange ink but piles it up. Two points of the flat sheet land on one point of the folded sheet, so the map is many-to-one. The folded view of a noisy image is then no longer a noisy image of the kind the model was trained on, and an estimate made for it has no unique way back to the two halves that produced it. Factorized Diffusion~\cite{geng2024factorized} and Diffusion Illusions~\cite{burgert2024diffusionillusions} handle views without an inverse, as sums of components or as stacked transparencies, but neither folds one sheet onto itself, where the same strokes must serve the flat drawing and, mirrored, the folded one. To our knowledge, illusions under such ink-accumulating folds have not been studied.

Sketches are where this question has an answer. A photograph has no empty paper, so two halves folded onto each other compete at every pixel and the result is a blend that reads as neither. A sketch leaves most of the sheet blank, so the strokes of two layers can pass between each other and still read as line art. But strokes make the problem hard in their own way. A stroke is a rigid curve, not a field of pixels. Whatever it does for the flat drawing it does, mirrored, for the folded one, so one set of control points must satisfy two drawings at once, and nothing can be moved for one reading without moving in the other. And because both readings pull on the same strokes, the stronger one tends to take them over while the other fades. The folded drawing also has only half the paper and gains nothing from ink that lands on ink, so the two halves must interleave rather than coincide, a constraint the flat drawing alone knows nothing about. And a sheet counts only if every reading works. A mean over readings hides a sheet that draws a perfect owl and no lighthouse.

\begin{figure*}[t]
    \centering
    \includegraphics[width=\textwidth]{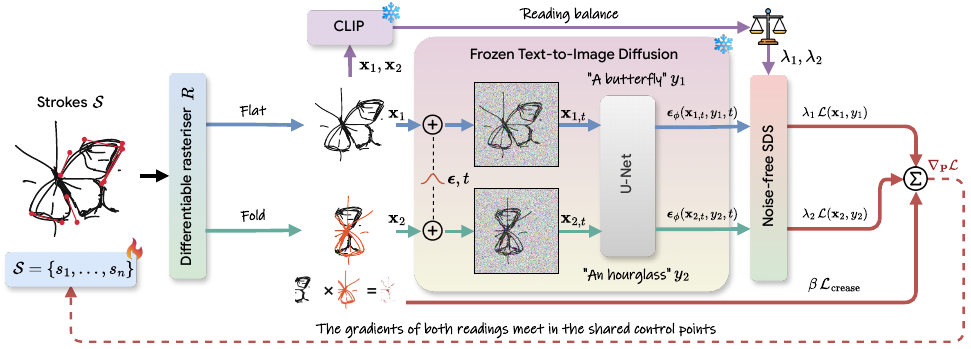}
    \caption{\textbf{Overview of \method.} The strokes are rendered once per step, and the image is folded for each reading (\cref{eq:fold}), with the resident strokes in black and the mirrored strokes in orange. Each reading is centred on a white square and scored against its prompt by a frozen diffusion model with noise-free score distillation, weighted by reading balance from CLIP, and a crease loss keeps the blurred layers of a fold apart. In shared-stroke distillation, the gradients of all readings meet in the shared control points. The sheet shown is a real result.}
    \label{fig:method}
\end{figure*}

\method meets these with four pieces (\cref{fig:method}). A composable fold operator maps strokes to strokes and agrees with the pixel
fold. It clips every stroke at the crease and reflects the moving half onto the resident one, so each reading is again a set of strokes, exportable as SVG and foldable again for the next crease. Because the operator is a reflection, the gradient of a folded reading reaches each stroke mirrored back to where it is drawn, and shared-stroke distillation updates every stroke from every reading it takes part in, with the sheet rendered once through a differentiable rasteriser~\cite{li2020diffvg} and each reading scored against its prompt with score distillation~\cite{poole2023dreamfusion}. Reading balance shifts weight towards the reading that CLIP finds weaker, so that one reading does not take over the shared strokes. A crease loss penalises ink of one layer landing on ink of the other, so the halves lay their strokes side by side and the folded drawing is built from both. FoldScore rates a sheet by its worst reading, so that our method, its ablations and its competitors are judged on whether every reading works.
On 50 pairs of concepts, \method scores 1.68 above the strongest vector sketch method in FoldScore (\cref{tab:main}). It scores level with Diffusion Illusions, which optimises a raster image through the same fold, while inking 6\% of the sheet against its 54\%.

Our contributions are as follows:
\begin{itemize}
\itemsep0em
\item To our knowledge, the first study of sketch illusions under folding, a reflection of the sheet onto itself that piles ink up instead of rearranging it, with an account of why sparse strokes, and not photographs, survive it.
\item A composable fold operator from strokes to strokes, shared-stroke distillation through which every stroke serves each reading it takes part in, reading balance, which shifts weight towards the weaker reading, and a crease loss that keeps the layers of a fold apart.
\item FoldScore, which rates a sheet by its worst reading on CLIP, ImageReward and HPS, so that methods are compared on whether every reading works.
\item Experiments on 50 pairs of concepts against two raster methods and four vector sketch methods, in which \method scores highest among the vector methods and level with the stronger raster one at about a
tenth of its ink.
\end{itemize}

\section{Related Work}
\label{sec:related}

\noindent\textbf{Multi-View Optical Illusions:}
Hybrid images~\cite{oliva2006hybrid} combine the low frequencies of one picture with the high frequencies of another, so that the reading changes with viewing distance. Diffusion Illusions~\cite{burgert2024diffusionillusions} optimises images through a fixed, differentiable arrangement with score distillation, including transparencies stacked over a backlight. Visual Anagrams~\cite{geng2024visualanagrams} denoises several views of one image and averages their noise estimates through the inverse views, which is exact for orthogonal views. Factorized Diffusion~\cite{geng2024factorized} conditions linear components of an image on different prompts. LookingGlass~\cite{chang2025lookingglass} generates anamorphoses, which reveal a second reading through a mirror or a lens. Ambigram generation~\cite{shirakawa2023ambigram} produces lettering that reads after a rotation. Later work~\cite{xu2025anagrammtl} keeps inverse views and names non-orthogonal transformations as open. All produce raster images, and none folds a sheet onto itself. In contrast, FoldSketch draws several readings by folding a single vector sketch under a many-to-one map, so that the same strokes must support every reading.

\noindent\textbf{Vector Sketch Generation:}
Differentiable rasterisation~\cite{li2020diffvg} lets vector graphics be optimised with losses on the rendered image. CLIPDraw~\cite{frans2022clipdraw} and CLIPasso~\cite{vinker2022clipasso} place strokes guided by CLIP~\cite{radford2021clip}, from a prompt and from a photograph respectively. VectorFusion~\cite{jain2023vectorfusion}, DiffSketcher~\cite{xing2023diffsketcher} and SVGDreamer~\cite{xing2024svgdreamer} use score distillation from a text-to-image diffusion model, and Word-As-Image~\cite{iluz2023wordasimage} deforms the outline of a letter to depict the meaning of a word. SketchAgent~\cite{vinker2025sketchagent} draws stroke by stroke with a multimodal language model. Each returns a single reading. Stroke of Surprise~\cite{cheng2026strokeofsurprise} is the closest to our goal. It optimises one set of B\'ezier strokes so that a prefix reads as one concept and the full set as another. Its second reading adds strokes over time, whereas ours reuses the same strokes in space. Unlike these methods, \method jointly optimises shared control points through flat and folded readings, allowing the same strokes to depict different concepts.

\noindent\textbf{Fold-Ins and Ambiguous Art:}
Shadow art~\cite{mitra2009shadowart} optimises a volume whose shadows in different directions form different images, and multi-view wire art~\cite{hsiao2018wireart} bends a wire sculpture to show a different line drawing from each viewpoint, which DreamWire~\cite{qu2024dreamwire} generates from text with score distillation. Both read their pictures through many-to-one projections, but optimise geometry in three dimensions rather than strokes on a sheet. Fold-ins~\cite{jaffee2011madfoldin} are drawn by hand, by an artist who predicts in the mind what the fold will show, a classic test of spatial imagery~\cite{shepard1972paperfolding}. The artist designs a page whose fold hides the middle and joins the outer strips into a second picture. \method turns a related craft into an optimisation problem, with see-through folds, several creases and text prompts in place of a hand-drawn page.

\section{Proposed Method}
\label{sec:proposed}
\noindent\textbf{Overview:}
\method takes $N$ prompts $y_1, \dots, y_N$ and a set of creases, and returns one set of strokes with the fold that reveals each reading (\cref{fig:method}). The first reading is the flat sheet. Each further reading is a fold sequence, one or more creases applied in turn, and shows the union of the resident and the mirrored strokes on the folded sheet. \cref{sec:background} recalls score distillation, \cref{sec:formulation} sets out how the layers of a fold combine and why sparse strokes survive them, and \cref{sec:foldsketch} gives the fold operator, the losses and FoldScore.

\subsection{Background}
\label{sec:background}
\noindent\textbf{Score Distillation:}
Score distillation sampling (SDS)~\cite{poole2023dreamfusion} uses a frozen diffusion model~\cite{ho2020ddpm} as a critic for any differentiable generator $\mathbf{x} = g(\theta)$. It noises an image $\mathbf{x}$, $\mathbf{x}_t = \sqrt{\bar{\alpha}_t}\,\mathbf{x} + \sqrt{1 - \bar{\alpha}_t}\,\boldsymbol{\epsilon}$ with $\boldsymbol{\epsilon} \sim \mathcal{N}(\mathbf{0}, \mathbf{I})$, and asks the network $\boldsymbol{\epsilon}_\phi(\mathbf{x}_t, y, t)$ to predict the noise given the prompt $y$ and the step $t$, and uses the prediction error as a gradient,
\begin{equation}
  \nabla_{\theta} \mathcal{L}_{\text{SDS}} = \mathbb{E}_{t, \boldsymbol{\epsilon}}\Bigl[ w(t)\,\bigl(\boldsymbol{\epsilon}_\phi(\mathbf{x}_t, y, t) - \boldsymbol{\epsilon}\bigr)\,\frac{\partial \mathbf{x}}{\partial \theta} \Bigr],
  \label{eq:sds}
\end{equation}
with a weight $w(t)$ per step, and with $\mathbf{x}$ replaced by its latent code $\mathcal{E}(\mathbf{x})$ in latent diffusion~\cite{rombach2022ldm}. When $g$ is a differentiable rasteriser~\cite{li2020diffvg}, $\theta$ holds the strokes' control points.

\subsection{Readings under Folding}
\label{sec:formulation}
\noindent\textbf{Sketch and Creases:}
A sketch $\mathcal{S} = \{s_i\}_{i=1}^{n}$ is a set of strokes on the sheet $\Omega = [-1, 1]^2$. Each stroke is a cubic B\'ezier curve with control points $\mathbf{P}_i \in \mathbb{R}^{4 \times 2}$, drawn in black with a fixed width, and a rasteriser $R$ turns a set of strokes into an image $I = R(\mathcal{S})$ whose pixels hold the ink coverage $I(p) \in [0, 1]$. We use two creases through the centre of the sheet. A fold along the vertical crease lays the left half over the right one, $f_v(x, y) = (|x|, y)$, and a fold along the horizontal crease lays the upper half over the lower one, $f_h(x, y) = (x, -|y|)$. A fold sequence $\mathbf{c} = (c_1, \dots, c_m)$ applies its creases in turn, $f_{\mathbf{c}} = f_{c_m} \circ \dots \circ f_{c_1}$, and the empty sequence leaves the sheet flat. Each crease sends two points of the sheet to one, so every fold is many-to-one.

\noindent\textbf{Ink Accumulation:}
A fold neither erases nor moves ink within a layer. Every point $q$ of the folded sheet shows the ink of all the points that land on it, and we assume that every layer stays visible, as on tracing paper or on a sheet held against the light. The folded image is then the union of the ink of all the layers that meet at $q$,
\begin{equation}
  I_{\mathbf{c}}(q) = 1 - \prod_{p \in f_{\mathbf{c}}^{-1}(q)} \bigl(1 - I(p)\bigr),
  \label{eq:fold}
\end{equation}
which for black ink is the product of the transmittances of the layers, the multiplicative blend that Diffusion Illusions~\cite{burgert2024diffusionillusions} uses for stacked transparencies. Away from the creases, $f_{\mathbf{c}}^{-1}(q)$ holds $2^m$ points for $m$ distinct perpendicular folds, one from each layer. We study the translucent sheet. The MAD fold-in, on opaque paper with two inward creases, joins the outer strips by translation instead.

\noindent\textbf{Problem:}
Given prompts $y_1, \dots, y_N$ and fold sequences $\mathbf{c}_1, \dots, \mathbf{c}_N$ with $\mathbf{c}_1$ empty, we seek strokes whose flat image reads as $y_1$ and whose folded images read as $y_2, \dots, y_N$,
\begin{equation}
  \mathcal{S}^{\ast} = \arg\min_{\mathcal{S}} \underbrace{\sum_{k=1}^{N} \lambda_k\, \mathcal{L}\bigl(I_{\mathbf{c}_k}, y_k\bigr)}_{\mathcal{L}_{\text{dist}}} + \beta\, \mathcal{L}_{\text{crease}}(\mathcal{S}),
  \label{eq:objective}
\end{equation}
where $I_{\mathbf{c}_k}$ is the image of $\mathcal{S}$ under $\mathbf{c}_k$ (\cref{eq:fold}), $\mathcal{L}$ the score distillation loss of a reading (\cref{eq:final}), and $\lambda_k$ and $\beta$ weigh the readings and the crease loss (\cref{eq:crease}). We evaluate the flat sheet and its vertical fold, and show three in \cref{fig:three}.

\noindent\textbf{Why There Is No Inverse:}
Visual Anagrams~\cite{geng2024visualanagrams} maps noise estimates back through the inverse views. A fold sends two points of the sheet to one point, so it has no inverse, and a folded noisy image is unlike any noisy image the model was trained on. On one layer, $f_{\mathbf{c}}$ is an identity, reflection or rotation by $\pi$, all isometries, so each stroke reaches every reading as a rigid copy and receives a well-defined gradient. We optimise strokes through the folds (\cref{sec:foldsketch}) instead of combining noise estimates.

\begin{figure}[H]
    \centering
    \includegraphics[width=\linewidth]{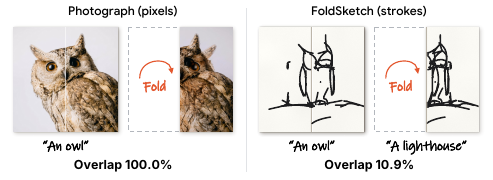}
    \vspace{-4mm}
    \caption{\textbf{Ink accumulation.} A photograph (left) and a \method sheet (right), each flat and folded along the vertical crease. Overlap is the share of the folded image's marked pixels (darker than paper) that receive marks from both layers. The sketch's strokes meet only where they cross and form a new drawing.}
    \label{fig:ink}
\end{figure}

\noindent\textbf{Why Sparse Strokes:}
In \cref{eq:fold}, the folded coverage lies between the largest coverage among the layers and their sum. If ink covers a fraction $\rho$ of the sheet, it covers at most $2^m \rho$ of the folded sheet, whose area is $2^m$ times smaller. A photograph has no empty paper, so every folded pixel mixes $2^m$ unrelated values. A sketch leaves most of the sheet blank, so its strokes can stay apart after a fold, and those that cross still read as lines (\cref{fig:ink}). In \cref{tab:main}, Diﬀusion Illusions inks half the sheet, our strokes 6\%.

\noindent\textbf{Why It Reads:}
A fold-in also leans on how the visual system groups strokes. Mirror symmetry is a classic grouping cue~\cite{wertheimer1923gestalt,wagemans2012gestalt}, and symmetry about a vertical axis, the axis of our crease, is detected faster than about a horizontal one~\cite{barlow1979symmetry,wagemans1995symmetry}, so we expect a flat drawing that is roughly symmetric about the crease, as
ours are, to read as one object, with the diﬀerences between its halves discounted. Those differences are the second drawing: a perfectly symmetric sheet folds onto itself and shows nothing new, and a sheet without symmetry hides nothing.  In five sheets each of \cref{fig:qualitative}, SketchAgent and Nano Banana Pro draw a symmetric picture, with 0.69 and 0.71 in mirror symmetry, the share of ink in one half within two pixels of ink in the mirrored other, and their folds score lower in CLIP than a single layer (\cref{tab:main}). DiffSketcher and CLIPasso score 0.21 and 0.28, and \method sits between the two groups at 0.52.

\subsection{FoldSketch}
\label{sec:foldsketch}
\noindent\textbf{Composable Fold Operator:}
\cref{eq:fold} folds the rendered image. The same fold can act on the strokes, so that every reading is itself a set of strokes that can be folded again. A crease $c$ splits the sheet into the half $\Omega_c^{+}$ that stays and the half $\Omega_c^{-}$ that moves, and $r_c$ is the reflection across $c$. We clip every stroke at the crease. With $\mathcal{S}_c^{\pm}$ the parts of the strokes in $\Omega_c^{\pm}$, the fold operator is defined as
\begin{equation}
  \mathcal{F}_c(\mathcal{S}) = \mathcal{S}_c^{+} \cup r_c\bigl(\mathcal{S}_c^{-}\bigr), \qquad \mathcal{F}_{\mathbf{c}} = \mathcal{F}_{c_m} \circ \dots \circ \mathcal{F}_{c_1}.
  \label{eq:operator}
\end{equation}
A reflection is affine, so it acts on a B\'ezier curve through its control points. Since black strokes are composited by their coverage, rendering the folded strokes reproduces \cref{eq:fold}, $R(\mathcal{F}_{\mathbf{c}}(\mathcal{S})) = I_{\mathbf{c}}$, up to anti-aliasing within a stroke width of a crease. We therefore render the sheet once per step and fold the image with \cref{eq:fold}, which gives every stroke the same gradient as $\mathcal{F}_{\mathbf{c}}$. The stroke-space operator exports the folded SVG, which matches the optimised folded image at ink IoU above 0.94 on every sheet we show.

\begin{figure*}[t]
    \centering
    \includegraphics[width=\textwidth]{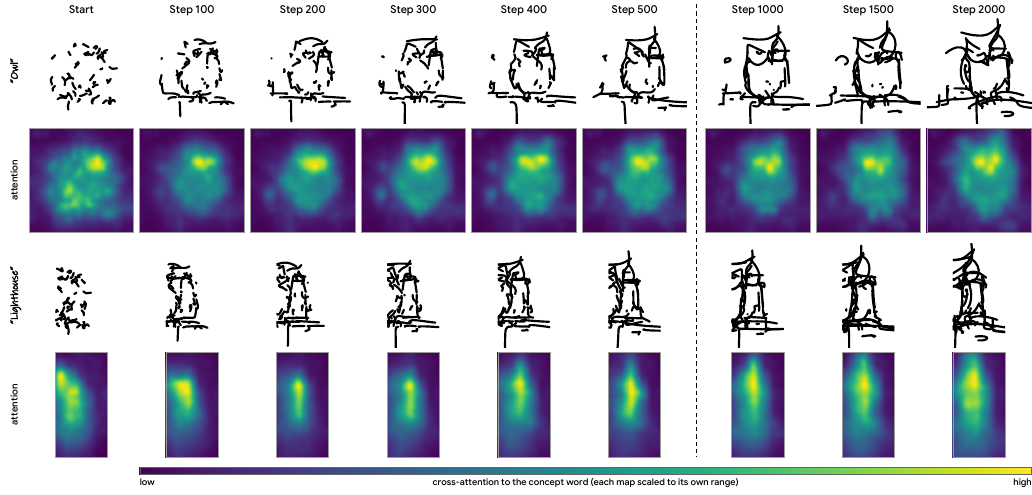}
    \caption{\textbf{Optimisation of one sheet.} An owl that folds into a lighthouse, at the start, every 100 steps up to step 500 and every 500 steps after the dashed line. The first and third rows show the strokes flat and folded along the vertical crease by the fold operator. The second and fourth rows show the cross-attention of Stable Diffusion to ``owl'' and ``lighthouse'', measured after training on that step's view at three noise levels and each scaled to its range. From step 100, ``owl'' attention peaks at the head, and from step 200 ``lighthouse'' attention runs along the tower. Each view shows the middle 720 of 1,024 pixels per side. The sheet is one of eight from extra runs, picked by eye.}
    \label{fig:progress}
\end{figure*}

\begingroup\clubpenalty=10000
\noindent\textbf{Shared-Stroke Distillation:}
The gradient of the distillation term $\mathcal{L}_{\text{dist}}$, the first term of \cref{eq:objective}, sums over the readings,
\begin{equation}
  \nabla_{\mathbf{P}} \mathcal{L}_{\text{dist}} = \sum_{k=1}^{N} \lambda_k\, \mathbb{E}_{t, \boldsymbol{\epsilon}}\Bigl[ w(t)\,\boldsymbol{\delta}_\phi(\mathbf{x}_{k,t}, y_k, t)\,\frac{\partial \mathbf{x}_k}{\partial \mathbf{P}} \Bigr],
  \label{eq:final}
\end{equation}
where each reading $\mathbf{x}_k = I_{\mathbf{c}_k}$ is centred on a white square, keeping its aspect ratio, and draws its own $\boldsymbol{\epsilon}$ and $t$. A stroke on a moving half appears in the flat reading where it is drawn and in a folded reading where the fold lays it. Its gradient sums both, the second reflected back through $r_c$, so one stroke learns to serve several drawings (\cref{fig:progress}).
Here $\boldsymbol{\delta}_\phi$ is the residual of noise-free score distillation~\cite{katzir2024nfsd}, in place of $\boldsymbol{\epsilon}_\phi - \boldsymbol{\epsilon}$ in \cref{eq:sds}, as in Stroke of Surprise~\cite{cheng2026strokeofsurprise}.\par\endgroup

\begingroup
\noindent\textbf{Reading Balance:}
With fixed weights, one reading often takes over the shared strokes while the other fades, which Xu \etal~\cite{xu2025anagrammtl} call concept domination. We therefore set $\lambda_k$ from how well each reads. Every 25 steps, CLIP~\cite{radford2021clip} with ViT-B/32 scores both readings against both prompts, and the margin $m_k$ of a reading, its own prompt's score minus the other's, is averaged with its last value. For two readings, with $r = \operatorname{clamp}(e^{(m_1 - m_2)/\tau}, 1/2, 2)$, the weights are
\begin{equation}
  \lambda_1 = \frac{2}{1 + r}, \qquad \lambda_2 = \frac{2r}{1 + r},
  \label{eq:balance}
\end{equation}
so the reading with the smaller margin gets up to twice the other's weight, and they average 1. The scale $\tau = 12.8$ gives a ratio of 1.5 at the median $|m_1 - m_2|$ of our fixed-weight sheets. CLIP gets no gradient, evaluation uses ViT-L/14, and ImageReward and HPS are never optimised.\par
\endgroup

\noindent\textbf{Crease Loss:}
A fold shows the union of its layers, so ink laid on ink adds nothing to the folded drawing. We adapt the overlay loss of Stroke of Surprise~\cite{cheng2026strokeofsurprise}, which keeps its later strokes off its earlier ones, to the layers of a fold, which are halves of the sheet rather than fixed sets of strokes. With $\mathcal{P}$ the pairs of layers $(\ell_i, \ell_j)$ of every folded reading, the ink maps of two halves placed in the folded frame, the crease loss is their mean blurred overlap,
\begin{equation}
  \mathcal{L}_{\text{crease}} = \frac{1}{|\mathcal{P}|\, HW} \sum_{(i,j) \in \mathcal{P}} \sum_{p} \tilde{\ell}_i(p)\, \tilde{\ell}_j(p),
  \label{eq:crease}
\end{equation}
where $\tilde{\ell} = G_\sigma * \ell$ is a layer blurred by a Gaussian kernel of width $\sigma$ and $HW$ the number of pixels of the sheet. It is the second term of \cref{eq:objective}. Its gradient pushes the two halves apart, so that each lays ink the other lacks. Without it, the two layers share 32.0\% of the smaller layer's ink instead of 23.7\% on our 50 pairs (\cref{sec:ablation}).

\noindent\textbf{FoldScore:}
FoldScore rates a sheet by its worst reading on three scores: CLIP score~\cite{radford2021clip}, which measures how well a reading matches its prompt, and ImageReward~\cite{xu2023imagereward} and HPS v2.1~\cite{wu2023hpsv2}, which are trained on human preferences between generated images. Each score $m \in \{\text{CLIP}, \text{IR}, \text{HPS}\}$ is standardised with a mean $\mu_m$ and a standard deviation $s_m$ fixed once on a reference pool of sheets, so that values compare across runs and methods,
\begin{equation}
  \mathrm{FoldScore}(\mathcal{S}) = \sum_{m} \frac{\min_k m(\mathbf{x}_k, y_k) - \mu_m}{s_m}.
  \label{eq:foldscore}
\end{equation}
The minimum over readings makes a sheet with one strong and one failed reading score low, which a mean would hide. The minimum follows the alignment score of Visual Anagrams~\cite{geng2024visualanagrams}, which also rates an illusion by its weakest view. FoldScore adds the two preference scores on one fixed scale. The top-1 sheet of a pair is the one with the highest FoldScore, for every method alike (\cref{sec:experiments}).

\section{Experiments}
\label{sec:experiments}
We compare \method with two raster and four vector sketch methods by eye and by score (\cref{sec:qualitative,sec:quantitative}) and test what each of the model parts contributes (\cref{sec:ablation}).

\noindent\textbf{Dataset:}
We evaluate on 50 pairs of concepts, each made of a flat concept and a folded one that is tall and narrow, such as a mushroom that folds into a floor lamp or a penguin that folds into a wine bottle. The sheet folds along the vertical crease, and each flat concept is roughly symmetric about it, so that half of its outline can serve the folded drawing. Every method makes four sheets per pair, one per seed. Two of the pairs, an owl that folds into a lighthouse and a face that folds into a giraffe, also recur as examples in our figures.

\noindent\textbf{Implementation Details:}
We build on the code of Stroke of Surprise~\cite{cheng2026strokeofsurprise}, with Stable Diffusion 2.1 base~\cite{rombach2022ldm} as the frozen prior and diffvg~\cite{li2020diffvg} as the rasteriser. Each sheet holds 48 cubic B\'ezier strokes, black on white, rendered at $1024 \times 1024$ pixels with lines 10 pixels wide. Strokes are randomly initialised in $[-0.4,0.4]^2$. Each prompt is the concept followed by \texttt{minimal 2d line drawing.}
\texttt{on a white background.} We optimise the control points with Adam for 2,000 steps, using a learning rate of 1.6, gradient clipping at 0.1, timesteps sampled uniformly from $[0.05,0.95]$ of the schedule, and guidance scale 100. The crease loss uses $\beta=5$ and $\sigma=4$ pixels. One sheet takes 9.1 minutes on an NVIDIA H200 (0.27 seconds per step, 7.5\,GB peak memory). The sheets in \cref{fig:teaser} were selected by eye among those whose readings CLIP ranks first, and some sheets of Figs. 1 and 5 used up to 5,000 steps, unlike the quantitative experiments.

\noindent\textbf{Evaluation Metrics:}
We evaluate the worst reading of each sheet using FoldScore (\cref{eq:foldscore}). CLIP score is $100\times$ the cosine similarity between the image and ``a drawing of'' the concept, using ViT-L/14. ImageReward (IR) and HPS v2.1 use ``a simple line drawing of'' the concept. Each reading is centred on a white square, and FoldScore is standardised using 156 reference sheets, frozen before any comparison. Recognition is the share of sheets whose every reading CLIP ranks first among the 100 distinct prompts of the 50 pairs. Coverage is the share of pairs with at least one recognised sheet, and ink is the share of flat-sheet pixels below 0.95 in luminance. Following Stroke of Surprise~\cite{cheng2026strokeofsurprise}, fold gain measures the folded reading's score minus that of its better individual layer. Semantic concealment follows Visual Anagrams~\cite{geng2024visualanagrams}, using bidirectional softmax-normalised CLIP similarities between readings and prompts.

\noindent\textbf{User Study:} Raters compared the top-1 sheets of \method and one competitor for each concept pair, shown flat and folded, and chose which better represented both prompts. Method names were hidden and presentation order randomised. Each rater completed 40 comparisons and four attention checks. Predefined exclusion criteria retained 25 of 30 raters, yielding 1,000 votes. Preference (Pref.) is the share of votes received by each method.

\noindent\textbf{Competitors:}
We compare with two raster methods, Diffusion Illusions~\cite{burgert2024diffusionillusions} and Nano Banana Pro~\cite{google2025nanobananapro}, and four vector sketch methods, CLIPasso~\cite{vinker2022clipasso}, SketchDreamer~\cite{qu2023sketchdreamer}, DiffSketcher~\cite{xing2023diffsketcher} and SketchAgent~\cite{vinker2025sketchagent}. The optimisation-based methods use their released code and losses, adapted to both readings through \cref{eq:fold}. Diffusion Illusions samples one reading per step and uses Stable Diffusion 1.5~\cite{rombach2022ldm}, with 4,000 score-distillation calls. CLIPasso and DiffSketcher use reference images from Stable Diffusion 1.5, cropped for the folded reading. SketchAgent and Nano Banana Pro are prompted in one or two requests, with the better protocol reported by top-1 FoldScore. SketchAgent uses Claude Sonnet 5 in place of its retired model. We exclude Visual Anagrams, which requires inverse views, and Stroke of Surprise, which does not support folded readings. Further settings are in the supplementary.

\subsection{Qualitative Evaluation}
\label{sec:qualitative}

\begin{figure}[t]
    \centering
    \includegraphics[width=\linewidth]{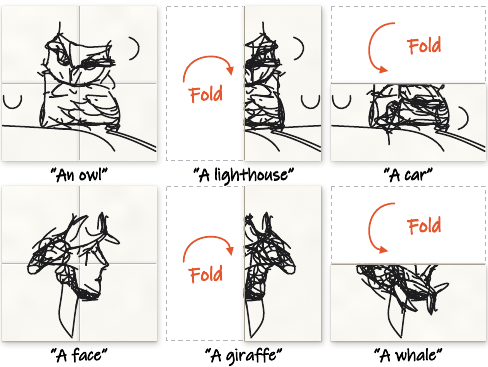}
    \vspace{-2mm}
    \caption{\textbf{Three readings.} Two sheets of 96 strokes, each flat, folded along the vertical crease and folded along the horizontal one. The dashed outline marks where the folded half lay.}
    \label{fig:three}
\end{figure}

\noindent\textbf{Three Readings:}
Shared-stroke distillation also supports three readings. With the flat sheet, its vertical fold and its horizontal fold weighted equally, we use 96 strokes to serve three drawings. \cref{fig:three} shows an owl, lighthouse and car, and a face, giraffe and whale. Across 32 sheets, CLIP ranks 68 of 96 readings first among 15 prompts, including every vertical fold. Eight sheets have all three readings recognised, with the best FoldScore of $+1.08$.

\begin{figure*}[t]
    \centering
    \includegraphics[width=\textwidth]{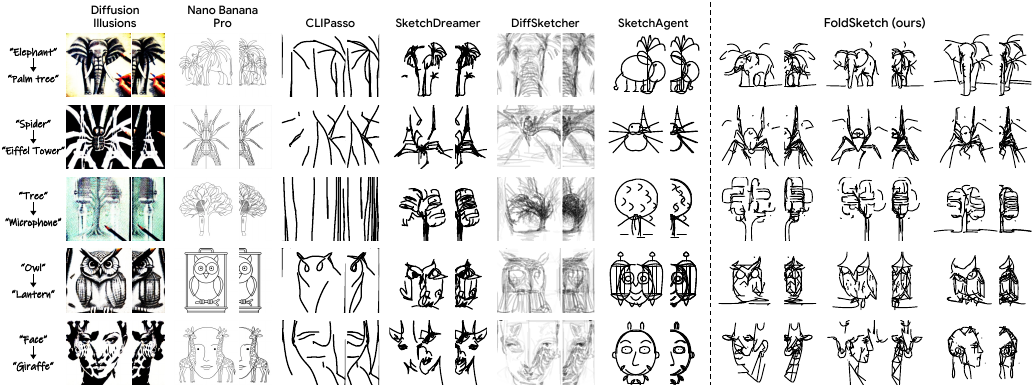}
    \caption{\textbf{Qualitative comparison} on five pairs, the flat concept above the arrow and the folded one below it, each sheet flat and folded, with three seeds of \method{}, which hides each half of the folded concept in the flat drawing.}
    \label{fig:qualitative}
\end{figure*}

\begin{table*}[t]
    \centering
    \caption{\textbf{Comparison} on 50 pairs folded along the vertical crease, four sheets per pair and method, with the metrics of \cref{sec:experiments}. Competitors show their top-1 sheet by FoldScore. Among the vector sketch top-1 rows, bold marks the best and underline the second.}
    \label{tab:main}
    \tablestyle{4.7pt}
    \begin{tabular}{l c ccc c c c c c}
        \toprule
        & \textbf{CLIP}\,$\uparrow$ & \multicolumn{3}{c}{\textbf{Fold gain}\,$\uparrow$} & \textbf{Concealment}\,$\uparrow$ & \textbf{Coverage}\,$\uparrow$ & \textbf{Ink}\,$\downarrow$ & \textbf{FoldScore}\,$\uparrow$ & \textbf{Pref.}\,$\uparrow$ \\
        \cmidrule(lr){3-5}
        \textbf{Method} & & \small CLIP & \small IR & \small HPS & & \small(\%) & \small(\%) & & \small(\%) \\
        \midrule
        \multicolumn{10}{l}{\textbf{\emph{Raster methods}}} \\
        Diffusion Illusions~\cite{burgert2024diffusionillusions} & 26.7 & $+3.1$ & $+0.72$ & $+0.028$ & 0.94 & 88.0 & 54.3 & $+2.57$ & 57.1 \\
        Nano Banana Pro~\cite{google2025nanobananapro} & 24.3 & $-0.9$ & $-0.17$ & $-0.004$ & 0.64 & 24.0 & 10.5 & $+2.37$ & 25.0 \\
        \hdashline
        \multicolumn{10}{l}{\textbf{\emph{Vector sketch methods}}} \\
        CLIPasso~\cite{vinker2022clipasso} & 23.3 & $+0.8$ & $+0.05$ & $-0.001$ & 0.70 & 0.0 & \underline{7.2} & $-2.91$ & 1.8 \\
        SketchDreamer~\cite{qu2023sketchdreamer} & \underline{26.2} & $\underline{+2.4}$ & $\underline{+0.56}$ & $\underline{+0.011}$ & \underline{0.88} & \underline{52.0} & 10.6 & $\underline{+1.00}$ & \underline{36.0} \\
        DiffSketcher~\cite{xing2023diffsketcher} & 22.3 & $+1.3$ & $+0.28$ & $-0.000$ & 0.82 & 20.0 & 34.5 & $-2.77$ & 17.0 \\
        SketchAgent~\cite{vinker2025sketchagent} & 23.7 & $-0.3$ & $+0.06$ & $-0.001$ & 0.70 & 10.0 & 7.9 & $+0.56$ & 19.5 \\
        \ours\textbf{\method (ours), all sheets} & \ours 25.3 & \ours $+3.0$ & \ours $+0.83$ & \ours $+0.011$ & \ours 0.87 & \ours 72.0 & \ours 5.7 & \ours $+1.16$ & \ours -- \\
        \ours\textbf{\method (ours), top-1} & \ours\textbf{26.8} & \ours $\mathbf{+4.2}$ & \ours $\mathbf{+1.03}$ & \ours $\mathbf{+0.016}$ & \ours\textbf{0.90} & \ours \textbf{72.0} & \ours\textbf{5.7} & \ours $\mathbf{+2.68}$ & \ours \textbf{73.9} \\
        \bottomrule
    \end{tabular}
    \vspace{-1mm}
\end{table*}

\noindent\textbf{Single Folds:}
\cref{fig:qualitative} shows the top-1 sheet of each competitor next to three sheets of \method. Competitors often favour one reading or reveal the folded concept on the flat sheet. SketchDreamer draws the folded concept into both halves, so its flat spider shows two Eiffel Towers. SketchAgent draws a symmetric picture whose fold is half of it, as for the tree and the face, or puts the folded concept beside the flat one, as for the owl between two lanterns. Nano Banana Pro also draws both concepts in plain view, a face between two giraffes or an Eiffel Tower inside a spider. CLIPasso places a few lines that outline at most the flat concept, and DiffSketcher shades the sheet with faint, dense strokes. Diffusion Illusions shows both concepts in most rows, but as shaded images that ink about half of the sheet, in colour for the elephant and the tree. In contrast, \method hides each half of the folded concept in the flat drawing, which shows an elephant, a spider or an owl, and the fold joins them into a palm tree, an Eiffel Tower or a lantern.

\subsection{Quantitative Evaluation}
\label{sec:quantitative}
\noindent\textbf{Readings:}
\method achieves the highest FoldScore among sketch methods (\cref{tab:main}). Its top-1 FoldScore of 2.68 exceeds SketchDreamer by 1.68, SketchAgent by 2.12, and DiffSketcher and CLIPasso by more than 5 across 50 pairs. Its best sheet scores higher than SketchDreamer's on 38 of the 50 pairs. It also achieves the highest CLIP score, fold gains and concealment among sketch methods. Diffusion Illusions scores comparably at 2.57, but inks 54\% of the sheet against \method's 6\%. Nano \mbox{Banana} Pro, with 10.5\% ink coverage, is comparable at 2.37 on the best sheet but 1.05 lower on the average of its four sheets. Its recognition coverage is only 24\%, compared with 72\% for \method.

\noindent\textbf{User Preference:} Raters preferred \method in 73.9\% of the comparisons, including at least 64\% against each competing sketch method and 75\% against Nano Banana Pro. SketchDreamer came closest at 36.0\%. Their ranking of the five sketch methods agrees with FoldScore. Diffusion Illusions, which inks half the sheet, received 57.1\% of 168 votes against \method, indicating no clear preference, consistent with their comparable FoldScores.

\subsection{Ablation Study}
\label{sec:ablation}
Each variant changes one component of \method and is evaluated on the same 50 pairs and four seeds as \cref{tab:main}, enabling paired comparisons. \cref{tab:ablation} evaluates the main contributions, while \cref{tab:design} examines design choices. We report 95\% bootstrap intervals over pairs.

\begin{table}[H]
    \centering
    \caption{\textbf{Contributions} on the 50 pairs of \cref{tab:main}. Flat and V are the mean CLIP scores of the two readings, FoldScore the mean over sheets, $\Delta$ its paired difference from the full method, and time minutes per sheet. Bold marks the best and underline the second.}
    \label{tab:ablation}
    \tablestyle{2.5pt}
    \begin{tabular}{l cccc c}
        \toprule
        \textbf{Variant} & \textbf{Flat}\,$\uparrow$ & \textbf{V}\,$\uparrow$ & \textbf{FoldScore}\,$\uparrow$ & $\Delta$ & \textbf{Time}\,$\downarrow$ \\
        \midrule
        w/o folded reading & \textbf{29.5} & 20.7 & $-2.57$ & $-3.73$ & \underline{9.2} \\
        w/o reading balance & 27.9 & 26.7 & $+0.71$ & $-0.45$ & 9.4 \\
        w/o crease loss & 27.9 & \underline{27.1} & $\underline{+1.08}$ & $-0.08$ & \textbf{9.1} \\
        \ours\textbf{\method (full)} & \ours \underline{28.0} & \ours \textbf{27.2} & \ours $\mathbf{+1.16}$ & \ours -- & \ours \textbf{9.1} \\
        \bottomrule
    \end{tabular}
    \vspace{-1mm}
\end{table}

\noindent\textbf{Shared-Stroke Distillation:}
Without the folded reading, only the flat drawing is optimised, increasing its CLIP score from 28.0 to 29.5 (\cref{tab:ablation}). However, the folded reading drops from 27.2 to 20.7, with only 4 of 200 sheets ranked first by CLIP against 115 for the full method. Recognition coverage falls from 36 to 2 of 50 pairs, and FoldScore decreases by 3.73 (95\% interval of 3.12 to 4.38). This confirms that meaningful folded drawings require joint optimisation of shared strokes through \cref{eq:fold}.

\begin{figure}[H]
    \centering
    \includegraphics[width=\linewidth]{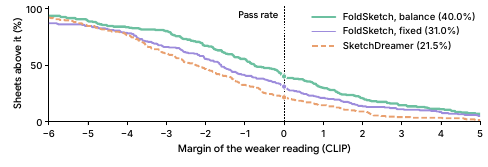}
    \vspace{-5mm}
    \caption{\textbf{Reading balance across margins.} The share of the 200 sheets of \cref{tab:main} whose weaker reading clears each CLIP margin, its own prompt's score minus the best other prompt's. At a margin of zero, each curve gives the share of sheets that pass recognition.}
    \label{fig:balance}
\end{figure}

\begingroup
\noindent\textbf{Reading Balance:}
On the 50 pairs of \cref{tab:main}, reading balance improves average FoldScore by 0.45 (95\% interval 0.14 to 0.76) over fixed weights, increasing sheet recognition from 31\% to 40\% (2.0 to 16.0 points more) and coverage from 64\% to 72\%. Both readings improve in CLIP score, particularly the folded one, from 26.7 to 27.2 (\cref{tab:ablation}). \cref{fig:balance} shows steady gains across weaker-reading margins from $-6$ to 5, outperforming SketchDreamer throughout.
\endgroup

\noindent\textbf{Crease Loss:}
Without the crease loss (\cref{eq:crease}), the folded layers share 32.0\% of the smaller layer's ink, compared with 23.7\% with the loss, counting pixels below 0.5 in luminance. Overlap increases in 48 of 50 pairs, by 8.3 points on average (95\% interval 6.8 to 9.7), with similar gaps at thresholds from 0.25 to 0.95. The loss thus encourages complementary stroke placement, while FoldScore stays level and runtime stays at 9.1 minutes per sheet with or without it (\cref{tab:ablation}). It removes about a quarter of the duplicated ink, which adds nothing to the folded drawing.

\begin{figure}[H]
    \centering
    \includegraphics[width=\linewidth]{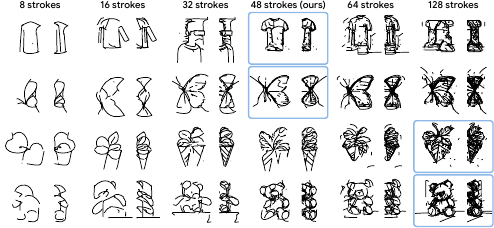}
    \vspace{-5mm}
    \caption{\textbf{Number of strokes.} Two pairs best at 48, two best at 128. A blue box marks the best sheet of each row by FoldScore.}
    \label{fig:ablation}
\end{figure}

\noindent\textbf{Number of Strokes:}
Each stroke serves both readings but also adds ink to the folded view. FoldScore improves rapidly up to 32 strokes before levelling off (\cref{tab:design}). Compared with 48 strokes, using 8 or 16 reduces FoldScore by 2.66 and 1.44, while 64 and 128 strokes yield comparable scores ($+0.11$ and $-0.19$) but increase ink coverage from 5.7\% to 6.7\% and 9.6\%. As observed in Stroke of Surprise~\cite{cheng2026strokeofsurprise}, the optimal budget depends on concept complexity: more strokes help detailed objects such as strawberries and teddy bears but clutter simpler outlines such as t-shirts and butterflies (\cref{fig:ablation}). We therefore retain 48 strokes for a better balance between semantic quality and ink coverage.

\noindent\textbf{Render Size:}
Rendering at 512 pixels, with lines, blur and learning rate scaled accordingly, reduces FoldScore by 0.56 (95\% interval 0.24 to 0.90), while lowering runtime from 9.1 to 6.3 minutes per sheet. We therefore retain 1024 pixels for better quality.

\begin{table}[H]
    \centering
    \caption{\textbf{Design choices} on the pairs and seeds of \cref{tab:ablation}, with the same scores and marks. \method draws 48 strokes rendered at 1024 pixels. The supplementary material gives further ablations.}
    \label{tab:design}
    \tablestyle{2.5pt}
    \begin{tabular}{l cccc c}
        \toprule
        \textbf{Variant} & \textbf{Flat}\,$\uparrow$ & \textbf{V}\,$\uparrow$ & \textbf{FoldScore}\,$\uparrow$ & $\Delta$ & \textbf{Time}\,$\downarrow$ \\
        \midrule
        8 strokes & 25.1 & 24.5 & $-1.50$ & $-2.66$ & 9.7 \\
        16 strokes & 26.7 & 26.3 & $-0.28$ & $-1.44$ & 9.7 \\
        32 strokes & 27.7 & 27.0 & $+0.94$ & $-0.22$ & 9.2 \\
        64 strokes & \underline{27.8} & \underline{27.1} & $\mathbf{+1.27}$ & $+0.11$ & 9.3 \\
        128 strokes & 27.7 & 26.9 & $+0.97$ & $-0.19$ & 9.5 \\
        512\,px render & 27.3 & 26.6 & $+0.60$ & $-0.56$ & \textbf{6.3} \\
        \ours\textbf{\method (full)} & \ours \textbf{28.0} & \ours \textbf{27.2} & \ours $\underline{+1.16}$ & \ours -- & \ours \underline{9.1} \\
        \bottomrule
    \end{tabular}
    \vspace{-1mm}
\end{table}

\section{Conclusion}
\label{sec:conclusion}
We asked whether the fold-in of the magazine page can be generated rather than drawn, in a stricter form. A fold piles ink up instead of rearranging it, so it is many-to-one and beyond denoising rules that need an inverse view, and only the sparseness of a sketch lets two folded layers read as one drawing. \method answers the question with a composable fold operator, shared-stroke distillation, reading balance and a crease loss. On 50 pairs of concepts it scores above every vector sketch method and level with a raster illusion that inks 54\% of the sheet against our 6\%. Rolling the paper and stacking cut pieces are other many-to-one maps, and
could reveal the next drawings a sheet can hold.

\section*{Acknowledgements}
We acknowledge the support of UK Research and Innovation (UKRI) through the UKRI AI: Centre for Doctoral Training in AI for Digital Media Inclusion under Grant No.: EP/Y030915/1. For the purpose of open access, the author has applied a Creative Commons Attribution (CC BY) license to any Author Accepted Manuscript version arising.

{
    \small
    \bibliographystyle{ieeenat_fullname}
    \bibliography{main}

@article{jastrow1899mindseye,
    author = {Joseph Jastrow},
    title = {{The Mind's Eye}},
    journal = {Popular Science Monthly},
    volume = {54},
    pages = {299--312},
    year = {1899}
}

@book{jaffee2011madfoldin,
    author = {Al Jaffee},
    title = {{The MAD Fold-In Collection: 1964--2010}},
    publisher = {Chronicle Books},
    year = {2011}
}

@article{oliva2006hybrid,
    author = {Aude Oliva and Antonio Torralba and Philippe G. Schyns},
    title = {{Hybrid images}},
    journal = {ACM Transactions on Graphics},
    volume = {25},
    number = {3},
    pages = {527--532},
    year = {2006}
}

@article{mitra2009shadowart,
    author = {Niloy J. Mitra and Mark Pauly},
    title = {{Shadow Art}},
    journal = {ACM Transactions on Graphics},
    volume = {28},
    number = {5},
    year = {2009}
}

@article{hsiao2018wireart,
    author = {Kai-Wen Hsiao and Jia-Bin Huang and Hung-Kuo Chu},
    title = {{Multi-view Wire Art}},
    journal = {ACM Transactions on Graphics},
    volume = {37},
    number = {6},
    year = {2018}
}

@inproceedings{shirakawa2023ambigram,
    author = {Takahiro Shirakawa and Seiichi Uchida},
    title = {{Ambigram Generation by A Diffusion Model}},
    booktitle = {ICDAR},
    year = {2023}
}

@inproceedings{geng2024visualanagrams,
    author = {Daniel Geng and Inbum Park and Andrew Owens},
    title = {{Visual Anagrams: Generating Multi-View Optical Illusions with Diffusion Models}},
    booktitle = {CVPR},
    year = {2024}
}

@inproceedings{geng2024factorized,
    author = {Daniel Geng and Inbum Park and Andrew Owens},
    title = {{Factorized Diffusion: Perceptual Illusions by Noise Decomposition}},
    booktitle = {ECCV},
    year = {2024}
}

@inproceedings{burgert2024diffusionillusions,
    author = {Ryan Burgert and Xiang Li and Abe Leite and Kanchana Ranasinghe and Michael S. Ryoo},
    title = {{Diffusion Illusions: Hiding Images in Plain Sight}},
    booktitle = {SIGGRAPH},
    year = {2024}
}

@inproceedings{chang2025lookingglass,
    author = {Pascal Chang and Sergio Sancho and Jingwei Tang and Markus Gross and Vinicius Azevedo},
    title = {{LookingGlass: Generative Anamorphoses via Laplacian Pyramid Warping}},
    booktitle = {CVPR},
    year = {2025}
}

@inproceedings{xu2025anagrammtl,
    author = {Zhiyuan Xu and Yinhe Chen and {Huan-ang} Gao and Weiyan Zhao and Guiyu Zhang and Hao Zhao},
    title = {{Diffusion-based Visual Anagram as Multi-task Learning}},
    booktitle = {WACV},
    year = {2025}
}

@inproceedings{qu2024dreamwire,
    author = {Zhiyu Qu and Lan Yang and Honggang Zhang and Tao Xiang and Kaiyue Pang and Yi-Zhe Song},
    title = {{Wired Perspectives: Multi-View Wire Art Embraces Generative AI}},
    booktitle = {CVPR},
    year = {2024}
}

@article{li2020diffvg,
    author = {Tzu-Mao Li and Michal Luk{\'a}{\v{c}} and Micha{\"e}l Gharbi and Jonathan Ragan-Kelley},
    title = {{Differentiable Vector Graphics Rasterization for Editing and Learning}},
    journal = {ACM Transactions on Graphics},
    volume = {39},
    number = {6},
    year = {2020}
}

@article{vinker2022clipasso,
    author = {Yael Vinker and Ehsan Pajouheshgar and Jessica Y. Bo and Roman Christian Bachmann and Amit Haim Bermano and Daniel Cohen-Or and Amir Zamir and Ariel Shamir},
    title = {{CLIPasso: Semantically-Aware Object Sketching}},
    journal = {ACM Transactions on Graphics},
    volume = {41},
    number = {4},
    year = {2022}
}

@inproceedings{frans2022clipdraw,
    author = {Kevin Frans and Lisa Soros and Olaf Witkowski},
    title = {{CLIPDraw: Exploring Text-to-Drawing Synthesis through Language-Image Encoders}},
    booktitle = {NeurIPS},
    year = {2022}
}

@inproceedings{jain2023vectorfusion,
    author = {Ajay Jain and Amber Xie and Pieter Abbeel},
    title = {{VectorFusion: Text-to-SVG by Abstracting Pixel-Based Diffusion Models}},
    booktitle = {CVPR},
    year = {2023}
}

@inproceedings{xing2023diffsketcher,
    author = {Ximing Xing and Chuang Wang and Haitao Zhou and Jing Zhang and Qian Yu and Dong Xu},
    title = {{DiffSketcher: Text Guided Vector Sketch Synthesis through Latent Diffusion Models}},
    booktitle = {NeurIPS},
    year = {2023}
}

@inproceedings{xing2024svgdreamer,
    author = {Ximing Xing and Haitao Zhou and Chuang Wang and Jing Zhang and Dong Xu and Qian Yu},
    title = {{SVGDreamer: Text Guided SVG Generation with Diffusion Model}},
    booktitle = {CVPR},
    year = {2024}
}

@inproceedings{vinker2025sketchagent,
    author = {Yael Vinker and Tamar {Rott Shaham} and Kristine Zheng and Alex Zhao and Judith E. Fan and Antonio Torralba},
    title = {{SketchAgent: Language-Driven Sequential Sketch Generation}},
    booktitle = {CVPR},
    year = {2025}
}

@article{iluz2023wordasimage,
    author = {Shir Iluz and Yael Vinker and Amir Hertz and Daniel Berio and Daniel Cohen-Or and Ariel Shamir},
    title = {{Word-As-Image for Semantic Typography}},
    journal = {ACM Transactions on Graphics},
    volume = {42},
    number = {4},
    year = {2023}
}

@inproceedings{cheng2026strokeofsurprise,
    author = {Huai-Hsun Cheng and Siang-Ling Zhang and Yu-Lun Liu},
    title = {{Stroke of Surprise: Progressive Semantic Illusions in Vector Sketching}},
    booktitle = {SIGGRAPH},
    year = {2026}
}

@inproceedings{qu2023sketchdreamer,
    author = {Zhiyu Qu and Tao Xiang and Yi-Zhe Song},
    title = {{SketchDreamer: Interactive Text-Augmented Creative Sketch Ideation}},
    booktitle = {BMVC},
    year = {2023}
}

@inproceedings{ho2020ddpm,
    author = {Jonathan Ho and Ajay Jain and Pieter Abbeel},
    title = {{Denoising Diffusion Probabilistic Models}},
    booktitle = {NeurIPS},
    year = {2020}
}

@inproceedings{rombach2022ldm,
    author = {Robin Rombach and Andreas Blattmann and Dominik Lorenz and Patrick Esser and Bj{\"o}rn Ommer},
    title = {{High-Resolution Image Synthesis with Latent Diffusion Models}},
    booktitle = {CVPR},
    year = {2022}
}

@inproceedings{poole2023dreamfusion,
    author = {Ben Poole and Ajay Jain and Jonathan T. Barron and Ben Mildenhall},
    title = {{DreamFusion: Text-to-3D using 2D Diffusion}},
    booktitle = {ICLR},
    year = {2023}
}

@inproceedings{katzir2024nfsd,
    author = {Oren Katzir and Or Patashnik and Daniel Cohen-Or and Dani Lischinski},
    title = {{Noise-free Score Distillation}},
    booktitle = {ICLR},
    year = {2024}
}

@inproceedings{radford2021clip,
    author = {Alec Radford and Jong Wook Kim and Chris Hallacy and Aditya Ramesh and Gabriel Goh and Sandhini Agarwal and Girish Sastry and Amanda Askell and Pamela Mishkin and Jack Clark and Gretchen Krueger and Ilya Sutskever},
    title = {{Learning Transferable Visual Models From Natural Language Supervision}},
    booktitle = {ICML},
    year = {2021}
}

@inproceedings{xu2023imagereward,
    author = {Jiazheng Xu and Xiao Liu and Yuchen Wu and Yuxuan Tong and Qinkai Li and Ming Ding and Jie Tang and Yuxiao Dong},
    title = {{ImageReward: Learning and Evaluating Human Preferences for Text-to-Image Generation}},
    booktitle = {NeurIPS},
    year = {2023}
}

@article{wu2023hpsv2,
      title={{Human Preference Score v2: A Solid Benchmark for Evaluating Human Preferences of Text-to-Image Synthesis}},
      author={Xiaoshi Wu and Yiming Hao and Keqiang Sun and Yixiong Chen and Feng Zhu and Rui Zhao and Hongsheng Li},
      year={2023},
      journal={arXiv preprint arXiv:2306.09341},
}

@misc{google2025nanobananapro,
    author = {{Google}},
    title = {{Nano Banana Pro (Gemini 3 Pro Image)}},
    howpublished = {\url{https://ai.google.dev/gemini-api/docs/models/gemini-3-pro-image}},
    year = {2025}
}

@article{shepard1972paperfolding,
    author = {Roger N. Shepard and Christine Feng},
    title = {{A chronometric study of mental paper folding}},
    journal = {Cognitive Psychology},
    volume = {3},
    number = {2},
    pages = {228--243},
    year = {1972}
}

@article{wertheimer1923gestalt,
    author = {Max Wertheimer},
    title = {{Untersuchungen zur Lehre von der Gestalt. II}},
    journal = {Psychologische Forschung},
    volume = {4},
    pages = {301--350},
    year = {1923}
}

@article{wagemans2012gestalt,
    author = {Johan Wagemans and James H. Elder and Michael Kubovy and Stephen E. Palmer and Mary A. Peterson and Manish Singh and R{\"u}diger von der Heydt},
    title = {{A century of Gestalt psychology in visual perception: I. Perceptual grouping and figure--ground organization}},
    journal = {Psychological Bulletin},
    volume = {138},
    number = {6},
    pages = {1172--1217},
    year = {2012}
}

@article{barlow1979symmetry,
    author = {H. B. Barlow and B. C. Reeves},
    title = {{The versatility and absolute efficiency of detecting mirror symmetry in random dot displays}},
    journal = {Vision Research},
    volume = {19},
    number = {7},
    pages = {783--793},
    year = {1979}
}

@article{wagemans1995symmetry,
    author = {Johan Wagemans},
    title = {{Detection of visual symmetries}},
    journal = {Spatial Vision},
    volume = {9},
    number = {1},
    pages = {9--32},
    year = {1995}
}

@inproceedings{zhang2018lpips,
  title     = {The Unreasonable Effectiveness of Deep Features as a Perceptual Metric},
  author    = {Zhang, Richard and Isola, Phillip and Efros, Alexei A. and Shechtman, Eli and Wang, Oliver},
  booktitle = {CVPR},
  year      = {2018}
}
}

\newpage
\twocolumn[
\begin{center}
    {\Large Supplementary Material for\\
    \textbf{FoldSketch: Drawings That Unfold When Folded}}
\end{center}
]
\renewcommand{\dbltopfraction}{0.98}
\renewcommand{\dblfloatpagefraction}{0.6}

\setcounter{section}{0}
\renewcommand{\thesection}{\Alph{section}}
\renewcommand{\theHsection}{supp.\Alph{section}}
\renewcommand{\theHfigure}{supp.\arabic{figure}}
\renewcommand{\theHtable}{supp.\arabic{table}}
\renewcommand{\theHequation}{supp.\arabic{equation}}

\setcounter{figure}{0}
\renewcommand{\figurename}{Fig.}
\renewcommand{\thefigure}{S\arabic{figure}}

\setcounter{table}{0}
\renewcommand{\tablename}{Table}
\renewcommand{\thetable}{S\arabic{table}}

\setcounter{equation}{0}
\renewcommand{\theequation}{S\arabic{equation}}

\section{Supplementary Material}
\label{sec:supplement}
\vspace{-1mm}

\noindent This supplementary material complements the main paper, ``FoldSketch: Drawings That Unfold When Folded'', with the full training recipe (\cref{sec:impl}), the evaluation protocol and the 50 concept pairs (\cref{sec:eval}), the settings of every competitor together with both prompting protocols of SketchAgent and Nano Banana Pro (\cref{sec:comp}), the design of the user study (\cref{sec:user}), further ablations (\cref{sec:abl}), further qualitative results with horizontal folds, proper nouns and scenes (\cref{sec:new}), a physical prototype on transparent film (\cref{sec:proto}), and the limitations of our approach with directions for future work (\cref{sec:limit}). Figures, tables, equations and sections of the main paper keep their numbers, as \cref{tab:main} or \cref{eq:balance}, and those of this document carry the prefix S, as \cref{tab:recipe} and \cref{fig:newpairs}.

\section{Implementation Details}
\label{sec:impl}

\noindent\textbf{Recipe:} \cref{tab:recipe} lists every setting of \method behind the quantitative results of the main paper. Each of the 50 pairs is run with seeds 0 to 3, which gives 200 sheets for every row of \cref{tab:main,tab:ablation,tab:design}.

\begin{table}[t]
\centering
\tablestyle{5pt}
\begin{tabular}{@{}lp{5.05cm}@{}}
\toprule
Setting & Value \\
\midrule
\multicolumn{2}{@{}l}{\textit{Sheet}} \\
Strokes & 48 cubic B\'ezier curves, one segment each \\
Ink and width & Black on white, lines 10 pixels wide \\
Canvas & $1024 \times 1024$ pixels, diffvg~\cite{li2020diffvg} \\
Initialisation & Control points at random in $[-0.4, 0.4]^2$ \\
\midrule
\multicolumn{2}{@{}l}{\textit{Prior}} \\
Model & Stable Diffusion 2.1 base~\cite{rombach2022ldm} \\
Gradient & Noise-free score distillation~\cite{katzir2024nfsd}, $w(t) = 1 - \bar{\alpha}_t$ \\
Guidance scale & 100 \\
Timesteps & Uniform in $[0.05, 0.95]$ of the schedule \\
Prompt & $\langle$concept$\rangle$ minimal 2d line drawing. on a white background. \\
\midrule
\multicolumn{2}{@{}l}{\textit{Optimisation}} \\
Optimiser & Adam on the control points, learning rate 1.6 \\
Steps & 2,000, gradient clipped at 0.1 \\
\midrule
\multicolumn{2}{@{}l}{\textit{Reading balance (\cref{eq:balance})}} \\
Scorer & CLIP ViT-B/32~\cite{radford2021clip}, ``a drawing of'' the concept \\
Updates & Every 25 steps, margins averaged with their last value \\
Ratio & Clamped to $[1/2, 2]$, scale $\tau = 12.77$ \\
\midrule
\multicolumn{2}{@{}l}{\textit{Crease loss (\cref{eq:crease})}} \\
Weight and blur & $\beta = 5$, Gaussian with $\sigma = 4$ pixels, kernel 31 \\
\midrule
\multicolumn{2}{@{}l}{\textit{Compute}} \\
Time per sheet & 9.1 minutes on one NVIDIA H200, 0.27 seconds per step \\
Peak memory & 7.5 GB \\
\bottomrule
\end{tabular}
\caption{\textbf{Training recipe.} Settings of \method for every quantitative result of the main paper. Coordinates are in sheet units, with the sheet spanning $[-1, 1]^2$, and pixels refer to the $1024 \times 1024$ canvas.}
\label{tab:recipe}
\end{table}

\noindent\textbf{Score Distillation:} Each reading is scored with noise-free score distillation~\cite{katzir2024nfsd}, as implemented by Stroke of Surprise~\cite{cheng2026strokeofsurprise}. In \cref{eq:final}, its residual takes the place of $\boldsymbol{\epsilon}_\phi - \boldsymbol{\epsilon}$ in \cref{eq:sds},
\begin{equation}
\boldsymbol{\delta}_\phi = s\,\bigl(\boldsymbol{\epsilon}_\phi(\mathbf{x}_t, y, t) - \boldsymbol{\epsilon}_\phi(\mathbf{x}_t, \varnothing, t)\bigr) + \boldsymbol{\delta}_D,
\label{eq:nfsd}
\end{equation}
where $s = 100$ is the guidance scale and $\varnothing$ the empty prompt. The second term is $\boldsymbol{\delta}_D = \boldsymbol{\epsilon}_\phi(\mathbf{x}_t, \varnothing, t)$ for $t < 200$ and $\boldsymbol{\epsilon}_\phi(\mathbf{x}_t, \varnothing, t) - \boldsymbol{\epsilon}_\phi(\mathbf{x}_t, y_{\text{neg}}, t)$ otherwise, with $t$ counted on the 1,000 steps of the schedule and $y_{\text{neg}}$ the default negative prompt of Stroke of Surprise. The weight is $w(t) = 1 - \bar{\alpha}_t$. Before it is scored, each reading is centred on a white square with its aspect ratio kept, so the folded view, the right half of $512 \times 1024$ pixels, sits in the middle of a $1024 \times 1024$ white image. Each reading draws its own noise and its own timestep at every step.

\noindent\textbf{Reading Balance:} Every 25 steps, 80 times in a run, CLIP ViT-B/32~\cite{radford2021clip} scores the flat reading $\mathbf{x}_1$ and the folded reading $\mathbf{x}_2$ against both prompts with the template ``a drawing of'', as 100 times the cosine similarity $c(\mathbf{x}, y)$. The margins are $m_1 = c(\mathbf{x}_1, y_1) - c(\mathbf{x}_1, y_2)$ and $m_2 = c(\mathbf{x}_2, y_2) - c(\mathbf{x}_2, y_1)$. Each is averaged with its previous value before \cref{eq:balance} sets the weights, which then stay fixed until the next update. At the first update of one sheet, for example, a flat margin of $-2.01$ and a folded margin of $+2.02$ give $\lambda_1 = 1.16$ and $\lambda_2 = 0.84$, so the weaker flat reading gains weight. The scale $\tau = 12.77$ is the median $|m_1 - m_2|$ of our fixed-weight sheets, 5.18, divided by $\ln 1.5$, so that the median sheet receives a ratio of 1.5, as in \cref{sec:foldsketch}. The weights scale only the distillation losses of the readings, and the crease loss keeps its fixed weight. CLIP runs without gradients, and none of the three models used for evaluation takes part in training at any point.

\noindent\textbf{Crease Loss:} For the vertical fold, the two layers of \cref{eq:crease} are the right half and the mirrored left half, both placed in the folded frame, and the ink of a layer is one minus its mean RGB value. Each ink map is blurred by a Gaussian with $\sigma = 4$ pixels and a kernel of 31 pixels, and $o_{ij}$ sums the product of the two blurred maps and divides it by the number of pixels of the whole sheet. The total loss is the weighted distillation loss plus $5\,\mathcal{L}_{\text{crease}}$. It adapts the overlay loss in the released code of Stroke of Surprise, which acts between its first strokes and the strokes it adds later.

\noindent\textbf{Compute:} One sheet takes 8.9 to 9.7 minutes on one NVIDIA H200, with a median of 9.1 minutes, 0.27 seconds per step and 7.5 GB of peak memory.

\section{Evaluation Details}
\label{sec:eval}

\noindent\textbf{Concept Pairs:} \cref{tab:pairs} lists the 50 pairs. Every flat concept is roughly symmetric about the vertical crease, and every folded concept is tall and narrow, so that it can fit the right half of the sheet. Two flat concepts, an owl and a face, appear in two pairs each, so the 50 pairs hold 98 distinct prompts in all, the pool used for recognition.

\begin{table*}[t]
\centering
\tablestyle{5pt}
\begin{tabular}{@{}rll@{\hspace{14pt}}rll@{\hspace{14pt}}rll@{}}
\toprule
No. & Flat & Folded & No. & Flat & Folded & No. & Flat & Folded \\
\midrule
1 & mushroom & floor lamp & 18$^*$ & owl & lighthouse & 35 & tooth & pair of pliers \\
2 & hot air balloon & light bulb & 19$^*$ & face & giraffe & 36 & chair & ladder \\
3 & pear & guitar & 20 & octopus & broom & 37 & leaf & feather \\
4 & face & vase & 21 & umbrella & saxophone & 38 & pair of glasses & key \\
5 & elephant & palm tree & 22 & teddy bear & traffic light & 39 & airplane & sword \\
6 & butterfly & hourglass & 23 & tent & lipstick & 40 & flying bat & axe \\
7 & snowman & fire hydrant & 24 & crab & cactus & 41 & alarm clock & tennis racquet \\
8 & penguin & wine bottle & 25 & sea turtle & violin & 42 & pair of headphones & hockey stick \\
9 & owl & lantern & 26 & lobster & rake & 43 & pair of binoculars & flashlight \\
10 & house & pencil & 27 & diamond & wine glass & 44 & open book & door \\
11 & spider & Eiffel Tower & 28 & pizza slice & crayon & 45 & rainbow & banana \\
12 & tree & microphone & 29 & cake & skyscraper & 46 & boomerang & golf club \\
13 & flower & lollipop & 30 & pineapple & asparagus spear & 47 & peanut & cello \\
14 & strawberry & ice cream & 31 & bee & syringe & 48 & moustache & flamingo \\
15 & rabbit & carrot & 32 & broccoli & paintbrush & 49 & chandelier & fork \\
16 & angel & candle & 33 & t-shirt & hammer & 50 & parachute & streetlight \\
17 & crown & castle & 34 & backpack & mailbox &  & &  \\
\bottomrule
\end{tabular}
\caption{\textbf{The 50 concept pairs} of \cref{sec:experiments}. Each sheet reads as the flat concept and, folded along the vertical crease, as the folded concept. Concepts are listed without their article, which the prompts keep, as in ``an owl'' or ``the Eiffel Tower''. $^*$ marks the two pairs that also appear as examples in the figures of the main paper.}
\label{tab:pairs}
\vspace{-2mm}
\end{table*}

\noindent\textbf{Scores:} CLIP score is 100 times the cosine similarity between a reading and ``a drawing of'' its concept, computed with CLIP ViT-L/14~\cite{radford2021clip}. ImageReward~\cite{xu2023imagereward} (v1.0) and HPS v2.1~\cite{wu2023hpsv2} score each reading against ``a simple line drawing of'' its concept. Every reading of every method is centred on a white square before scoring, as in training. FoldScore (\cref{eq:foldscore}) standardises each score with constants fixed once, on 156 sheets from our pilot runs, before any experiment of \cref{sec:experiments}, and never changed afterwards: $\mu = 24.017$ and $s = 2.922$ for CLIP, $\mu = -1.257$ and $s = 0.844$ for ImageReward, and $\mu = 0.188$ and $s = 0.015$ for HPS.

\noindent\textbf{Recognition and Coverage:} A reading is recognised when CLIP ranks its own prompt first among the 98 distinct prompts of the 50 pairs, and a sheet passes when every reading is recognised. Ranking among all 98 prompts is stricter than choosing between the two prompts of a pair, since each reading must also beat the concepts of the other pairs. Pass rate is the share of sheets that pass, and coverage is the share of pairs with at least one passing sheet among their four.

\noindent\textbf{Ink, Fold Gain and Concealment:} Ink is the share of pixels of the flat sheet whose luminance, $0.299R + 0.587G + 0.114B$ on a scale from 0 to 1, is below 0.95. Fold gain compares the folded reading with its two layers, each half drawn alone in the folded frame and padded in the same way. It is the folded reading's score minus the higher of the two layers' scores, computed for CLIP, ImageReward and HPS alike, so it is positive only when the folded drawing reads better than either half on its own. Concealment follows Visual Anagrams~\cite{geng2024visualanagrams}. With $S$ the $2 \times 2$ matrix of CLIP similarities between the two readings and the two prompts, multiplied by 100, it is the mean diagonal entry of the softmax of $S$ taken over prompts and over readings, averaged over the two directions. It is low when a reading matches the other reading's prompt, for example when the flat drawing already shows the folded concept.

\noindent\textbf{Selection and Intervals:} For every method, the top-1 sheet of a pair is the one of its four sheets with the highest FoldScore, and rows over all sheets average the 200 sheets. Intervals are 95\% percentile bootstrap intervals over the 50 pairs with 10,000 resamples. A paired difference first takes each method's mean over the four sheets of a pair, or its top-1 sheet, and then resamples pairs, drawing the same pairs for both methods. The user study resamples raters instead (\cref{sec:user}). Every resampling uses one fixed random seed, so every interval can be reproduced.

\section{Competitor Settings}
\label{sec:comp}

\noindent\textbf{Shared Protocol:} Every competitor runs its released code at a fixed commit (\cref{tab:comp}) with its own defaults for strokes, initialisation, loss, optimiser, schedule, prompt form and output, apart from the changes given below. Each makes four sheets per pair, with seeds 0 to 3 for the optimisation methods and four independent requests for the two methods we query through an API, which offers no seed. We fold every returned sheet with \cref{eq:fold} and score it exactly as we score \method. Competitor sheets are saved at $512 \times 512$ pixels, the vector ones rendered again from their strokes with widths scaled to match. The optimisation methods give each reading the method's own loss, with equal weights.

\begin{table*}[t]
\centering
\tablestyle{5pt}
\begin{tabularx}{\textwidth}{@{}lllll>{\raggedright\arraybackslash}X@{}}
\toprule
Method & Code & Model & Sheet & Budget & Prompt \\
\midrule
\multicolumn{6}{@{}l}{\textit{Raster methods}} \\
Diffusion Illusions~\cite{burgert2024diffusionillusions} & 71646f32 & SD 1.5 & Fourier-feature image, $256^2$ & 4,000 iterations & ``an intricate detailed hb pencil sketch of'' $\langle c\rangle$ \\
Nano Banana Pro~\cite{google2025nanobananapro} & API & gemini-3-pro-image & Image, 1K, square & 1 or 2 requests & \cref{tab:prompts} \\
\midrule
\multicolumn{6}{@{}l}{\textit{Vector sketch methods}} \\
CLIPasso~\cite{vinker2022clipasso} & 92262f70 & CLIP RN101 & 16 strokes, 224 pixels & 2,001 iterations & References: ``a photo of $\langle c\rangle$, on a white background'' \\
SketchDreamer~\cite{qu2023sketchdreamer} & 6be09bd5 & SD 1.5 & 16 strokes, 512 pixels & 1,001 iterations & ``a simple drawing of $\langle c\rangle$ on a white background.'' \\
DiffSketcher~\cite{xing2023diffsketcher} & 0319b290 & CLIP RN101, LPIPS & 128 strokes, 224 pixels & 2,000 iterations & Targets: ``a photo of $\langle c\rangle$'' \\
SketchAgent~\cite{vinker2025sketchagent} & 14c40430 & Claude Sonnet 5 & $50 \times 50$ grid & 1 or 2 requests & Its own prompts with the fold added \\
\midrule
\ours \method (ours) & \ours -- & \ours SD 2.1 base & \ours 48 strokes, 1024 pixels & \ours 2,000 steps & \ours $\langle c\rangle$ ``minimal 2d line drawing. on a white background.'' \\
\bottomrule
\end{tabularx}
\caption{\textbf{Competitor settings.} Code is the commit of each method's released repository. Model is the network that guides the sheet. SD is Stable Diffusion. Budget counts optimisation steps or requests per sheet, and $\langle c\rangle$ stands for the concept. CLIPasso and DiffSketcher draw towards reference or target images from Stable Diffusion 1.5, made with the prompts shown.}
\label{tab:comp}
\vspace{-2mm}
\end{table*}

\noindent\textbf{Diffusion Illusions:} We use its two-view notebook with the fold in place of the rotation: the flat view is the image itself, and the folded view is \cref{eq:fold} applied to it and padded to a square. The image is a learnable Fourier-feature texture of $256 \times 256$ pixels, upsampled to 512 inside each step and trained with SGD at a learning rate of $10^{-4}$, one randomly chosen view per iteration, guidance 60, timesteps from 10 to 990 and an empty negative prompt, as in the notebook. We run it for 4,000 iterations, the distillation calls of \method (2,000 steps for each of two readings), and use Stable Diffusion 1.5 in place of 1.4, which has the same architecture. The texture has three colour channels, as released, and the sheet is the image after the last iteration.

\noindent\textbf{Nano Banana Pro:} We call gemini-3-pro-image with image output only, aspect ratio 1:1 and its default size of 1K. Its thinking cannot be turned off, and the last image of the reply that is not part of a thought is the sheet. \cref{tab:prompts} gives the prompts. The image is cropped to a centred square if needed and resized to $512 \times 512$ pixels. Some images came with a thin frame along the border, which would count as ink, so any edge row or column within 3 pixels of the border whose mean luminance is below 0.75 is set to white. We keep the original replies.

\begin{table*}[t]
\centering
\tablestyle{5pt}
\begin{tabular}{@{}lp{14.6cm}@{}}
\toprule
Request & Prompt \\
\midrule
One request & Make a square drawing that is a fold illusion. The square sheet will be folded in half along its vertical centre line: the left half is folded over onto the right half, like closing a book. The paper is translucent, so the folded view shows the lines of the right half together with the lines of the left half, mirrored left to right. Unfolded, the drawing must look like $\langle A\rangle$. Folded, the visible half must look like $\langle B\rangle$. Show only the unfolded square sheet, seen flat from the front. $\langle$Style$\rangle$ \\
\midrule
First of two & Make a square drawing of $\langle A\rangle$. $\langle$Style$\rangle$ \\
\midrule
Second of two & The first image is a square sheet with a line drawing of $\langle A\rangle$. The square sheet will be folded in half along its vertical centre line: the left half is folded over onto the right half, like closing a book. The paper is translucent, so the folded view shows the lines of the right half together with the lines of the left half, mirrored left to right. The second image shows the folded view of this sheet as it is now. Edit the first image by adding lines so that the folded view becomes $\langle B\rangle$, while the unfolded sheet still looks like $\langle A\rangle$. Keep the existing lines. Show only the edited unfolded square sheet, seen flat from the front. $\langle$Style$\rangle$ \\
\midrule
$\langle$Style$\rangle$ & Style: a minimal 2D line drawing in thin black lines on a plain white background, with no shading, no colour, no text, no border and no fold line. The drawing itself fills the whole image: do not show a photo of a sheet of paper, its edges or its shadow. \\
\bottomrule
\end{tabular}
\caption{\textbf{Prompts of Nano Banana Pro} for the vertical fold, word for word, with $\langle A\rangle$ the flat concept and $\langle B\rangle$ the folded concept. The second of two requests also attaches the sheet and its folded view, made by our fold operator, as the two images it mentions.}
\label{tab:prompts}
\vspace{-2mm}
\end{table*}

\noindent\textbf{CLIPasso:} CLIPasso draws from an image, so each reading gets a reference image from Stable Diffusion 1.5 with the prompt of \cref{tab:comp}, 100 denoising steps, guidance 7.5 and the sheet's seed. The folded reading can only show the right half of the sheet, so its reference is the central strip of its image, padded white to a square like the folded view. Its 16 strokes, 1.5 pixels wide on a 224-pixel canvas, start from its saliency-based initialisation, 8 on each reading's reference, with the folded reading's starts placed alternately on the right half and mirrored onto the left half. The loss is its CLIP RN101 distance summed over the two readings, optimised with Adam at a learning rate of 1.0. As its launcher does, we keep the sheet with its lowest evaluation loss and use its early stop.

\noindent\textbf{SketchDreamer:} It optimises 16 strokes of five cubic segments each, 3 pixels wide on a 512-pixel canvas and starting at random in the central $[0.2, 0.8]^2$ of the canvas, with Adam at a learning rate of 1.0 for its 1,001 iterations. Its score distillation uses Stable Diffusion 1.5 with guidance 100 and timesteps from 50 to 950, one step per reading in every iteration. Strokes shorter than 100 pixels are redrawn every 50 iterations, as in its code, and we keep its final sheet.

\noindent\textbf{DiffSketcher:} For each reading, its Stable Diffusion 1.5 pipeline makes a target image with the prompt of \cref{tab:comp} and records cross-attention. The attention to the concept's last token, weighted by an edge map of the target, places the starts of its 128 strokes, 64 per reading, which are 1.5 pixels wide on a 224-pixel canvas. As for CLIPasso, the folded reading's target is the central strip of its image padded white, and its starts are placed on both halves. The loss is its CLIP RN101 distance plus 0.2 times LPIPS~\cite{zhang2018lpips} to each reading's target, summed over the readings, for 2,000 iterations, and we keep its best sheet by CLIP distance, evaluated every 10 iterations as in its released command.

\noindent\textbf{SketchAgent:} Its default model, Claude 3.5 Sonnet, has been retired. We use Claude Sonnet 5 (claude-sonnet-5) with thinking off, so that the model reasons in the tags its prompts ask for, as the original model did, and we keep its prompts, stroke format, B\'ezier fitting and three attempts per request. The fold lengthens this reasoning, and in a first test all six fold requests ran out of its 3,000 output tokens before the first stroke, so we raise the limit to 16,000. We sample with the default settings of the API, since its deterministic mode would give the same sheet for every seed. The fold is stated on its grid: the crease lies between columns 25 and 26, and a point in column $x$ of the left half lands in column $51 - x$. Its strokes are cropped to its drawing area of 600 pixels and rendered on a 512-pixel sheet, with their width scaled from 7 to 5.97 pixels.

\noindent\textbf{Two Protocols:} Neither SketchAgent nor Nano Banana Pro optimises through the fold, so the fold enters only through the request. In the one-request protocol, a single request explains the fold and asks for a sheet that reads as the flat concept unfolded and as the folded concept folded. In the two-request protocol, the first request asks for the flat concept alone, with SketchAgent's own prompt or a plain drawing request for Nano Banana Pro. The second request shows the sheet and its folded view, made by our fold operator, and asks for additions that turn the folded view into the folded concept while the sheet keeps the flat one. SketchAgent adds strokes to its first ones, and Nano Banana Pro edits its image. Before any run, we fixed that \cref{tab:main} reports, for each method, the protocol with the higher top-1 FoldScore. \cref{tab:protocols} gives both. Against the protocols left out of \cref{tab:main}, the top-1 FoldScore of \method is higher by 1.58 (0.82 to 2.35) for Nano Banana Pro with one request and by 2.15 (1.37 to 2.93) for SketchAgent with two.

\begin{table*}[t]
\centering
\tablestyle{5pt}
\begin{tabular}{@{}llcccccccc@{}}
\toprule
& & & \multicolumn{3}{c}{Fold gain $\uparrow$} & & & & \\
\cmidrule(lr){4-6}
Method & Requests & CLIP $\uparrow$ & CLIP & IR & HPS & Concealment $\uparrow$ & Coverage (\%) $\uparrow$ & Ink (\%) $\downarrow$ & FoldScore $\uparrow$ \\
\midrule
Nano Banana Pro & One & 23.4 & $-0.6$ & $-0.28$ & $-0.003$ & 0.61 & 20.0 & 13.8 & $+1.10$ \\
Nano Banana Pro & Two (\cref{tab:main}) & 24.3 & $-0.9$ & $-0.17$ & $-0.004$ & 0.64 & 24.0 & 10.5 & $+2.37$ \\
\midrule
SketchAgent & One (\cref{tab:main}) & 23.7 & $-0.3$ & $+0.06$ & $-0.001$ & 0.70 & 10.0 & 7.9 & $+0.56$ \\
SketchAgent & Two & 24.3 & $-0.1$ & $+0.08$ & $-0.002$ & 0.74 & 14.0 & 8.3 & $+0.53$ \\
\midrule
\ours \method (ours) & \ours -- & \ours \textbf{26.8} & \ours $\mathbf{+4.2}$ & \ours $\mathbf{+1.03}$ & \ours $\mathbf{+0.016}$ & \ours \textbf{0.90} & \ours \textbf{72.0} & \ours \textbf{5.7} & \ours $\mathbf{+2.68}$ \\
\bottomrule
\end{tabular}
\caption{\textbf{Both prompting protocols} of Nano Banana Pro and SketchAgent on the 50 pairs, with the metrics and top-1 sheets of \cref{tab:main}. Requests counts the requests per sheet, and the protocols marked \cref{tab:main} are the ones reported there, chosen by top-1 FoldScore by a rule fixed before any run. The last row repeats the top-1 row of \method. Bold marks the best value in each column.}
\label{tab:protocols}
\vspace{-2mm}
\end{table*}

\section{User Study}
\label{sec:user}

\noindent\textbf{Design:} The study compares \method with each competitor of \cref{tab:main} on all 50 pairs, using the top-1 sheet of every method, the same sheets as the top-1 rows of \cref{tab:main}. This gives 300 comparisons, one per pair and competitor, each planned for four raters. They are split into 30 blocks of 40 comparisons, each with 40 different pairs and six or seven comparisons per competitor, plus four attention checks. An attention check shows the sheet of the pair named in the question next to the same method's sheet for another pair, so that it has one right answer.

\noindent\textbf{Interface:} An introduction explains that each sheet is drawn on see-through paper, which shows one thing flat and a second thing when the left half is folded over the right, with a diagram of the fold. Each screen then shows Sheet A and Sheet B, each flat as a square and, beside it, folded as its right half at the same height, and asks ``Which sheet is better? Flat it should look like $\langle A\rangle$, folded it should look like $\langle B\rangle$.'' Method names never reach the page, and the side that shows \method is drawn at random for every rater and comparison. The study takes about ten minutes, is voluntary and anonymous, asks raters to confirm that they are 18 or older, collects no personal information and records only the choices and the time they take.

\noindent\textbf{Exclusion and Scoring:} We fixed the rules before any answer arrived. A rater is excluded after failing at least two of the four attention checks or answering in a median time below 2.0 seconds. Of 30 raters, each answering one block, 25 are kept, and all five excluded raters failed at least two checks. The kept raters took a median of 7.0 seconds per answer, from 3.6 to 13.4 across raters, and their 1,000 votes cover all 300 comparisons. The Pref. of a competitor is the share of votes for it in its comparisons with \method, and the Pref. of \method is its share of all 1,000 votes. Intervals are 95\% bootstrap intervals over raters with 10,000 resamples. \cref{tab:user} breaks the votes down by competitor. \method is preferred over every sketch method, from 64.0\% of the votes against SketchDreamer to 98.2\% against CLIPasso, and over Nano Banana Pro in 75.0\% of them.

\begin{table}[H]
\centering
\tablestyle{5pt}
\begin{tabular}{@{}lccc@{}}
\toprule
Competitor & Votes & Competitor (\%) & Ours (\%) \\
\midrule
Diffusion Illusions & 168 & 57.1 (45.6 to 68.3) & 42.9 \\
Nano Banana Pro & 168 & 25.0 (16.5 to 34.9) & 75.0 \\
CLIPasso & 166 & 1.8 (0.0 to 3.7) & 98.2 \\
SketchDreamer & 164 & 36.0 (28.4 to 43.6) & 64.0 \\
DiffSketcher & 165 & 17.0 (10.2 to 24.1) & 83.0 \\
SketchAgent & 169 & 19.5 (13.1 to 26.3) & 80.5 \\
\midrule
\ours All & \ours 1,000 & \ours -- & \ours 73.9 \\
\bottomrule
\end{tabular}
\caption{\textbf{User study by competitor.} Votes in the comparisons of each competitor with \method on the 50 pairs, the share for each side, with ours for \method, and in brackets 95\% bootstrap intervals over the 25 kept raters. Over all 1,000 votes, \method receives 73.9\% (69.9 to 77.7). Competitor shares are the Pref. column of \cref{tab:main}.}
\label{tab:user}
\end{table}

\begin{figure*}[tp]
\centering
\includegraphics[scale=0.95]{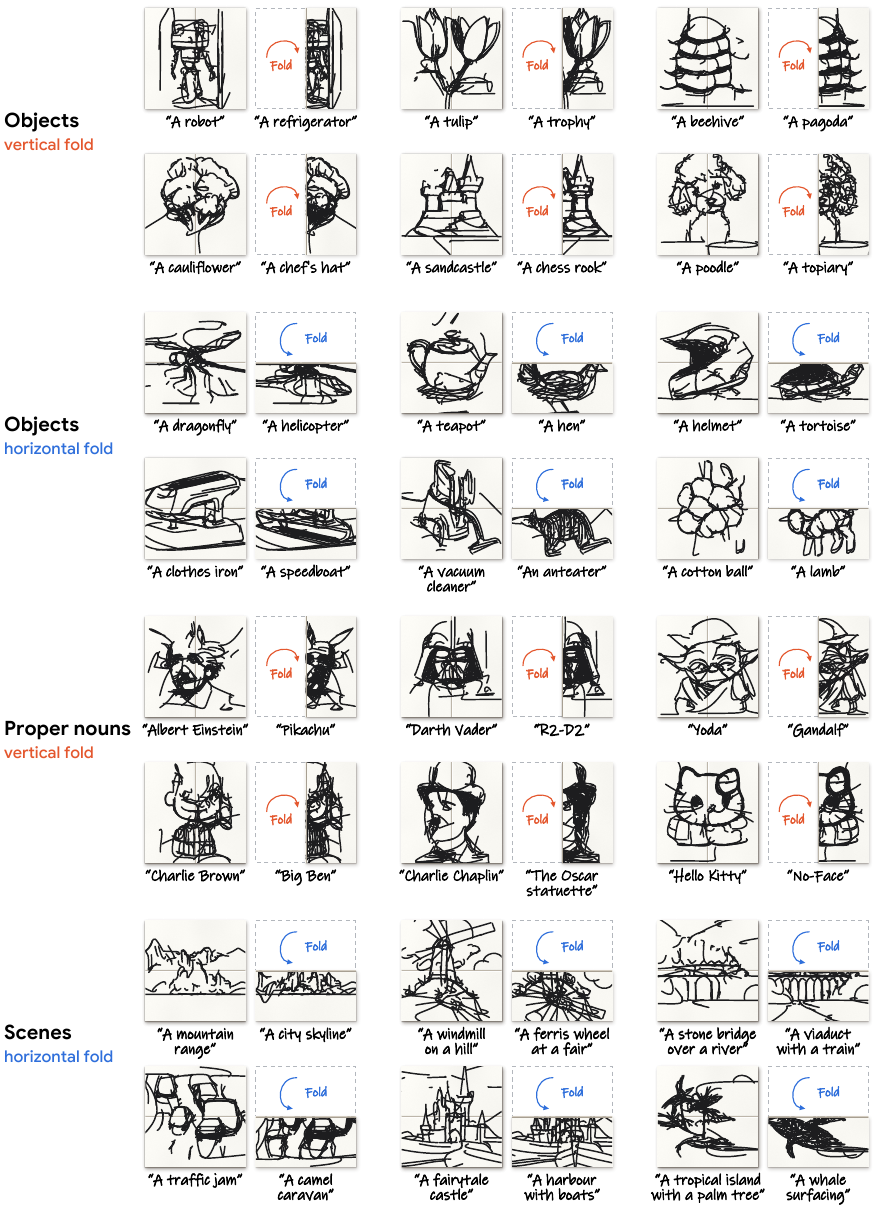}
\caption{\textbf{FoldSketch on new pairs.} Six sheets for each kind of pair, each flat and then folded, with 128 strokes after 5,000 steps. The vertical fold (orange) lays the left half over the right one, the horizontal fold (blue) lays the upper half over the lower one, and the dashed outline marks where the folded half lay. For display, strokes are drawn 1.5 times thicker, each view is cropped to 80\% of the sheet's side, and short strokes that lie apart from all others are hidden.}
\label{fig:newpairs}
\end{figure*}

\begin{figure*}[p]
\centering
\includegraphics[width=\textwidth]{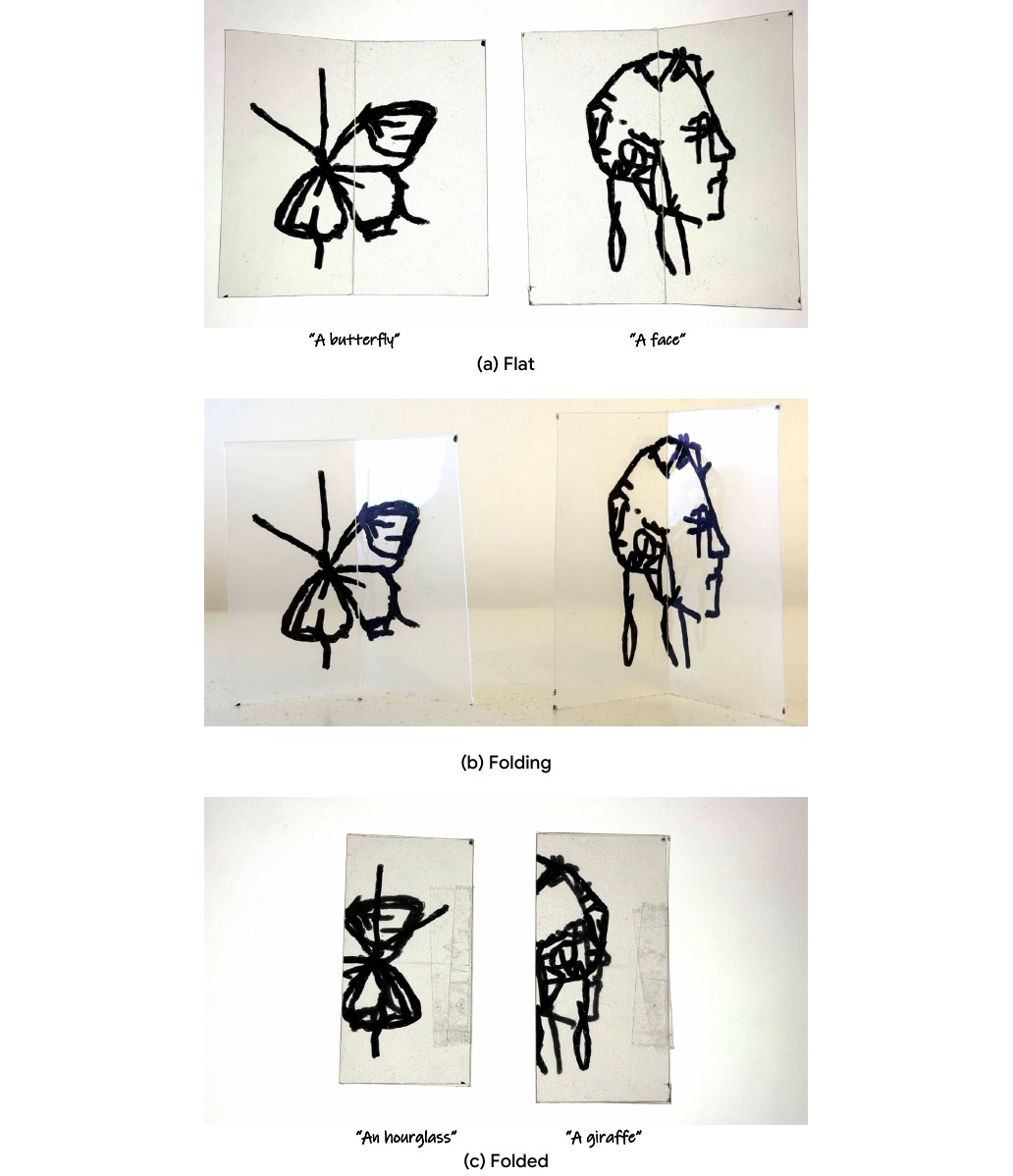}
\caption{\textbf{Physical prototype.} The butterfly and the face of \cref{fig:teaser}, redrawn by hand with a permanent marker on transparent plastic film. (a)~Flat, the two sheets show a butterfly and a face. (b)~Each sheet is folded along its vertical centre line, the left half over the right one. (c)~Folded, they show an hourglass and a giraffe.}
\label{fig:proto}
\end{figure*}

\section{Further Ablations}
\label{sec:abl}

\noindent Every row below uses the 50 pairs and seeds 0 to 3 of \cref{tab:main} and changes one setting of \method, as in \cref{tab:ablation}.

\noindent\textbf{Readings under Each Contribution:} \cref{tab:abl_read} counts, for every row of \cref{tab:ablation}, the sheets on which each reading is recognised. Without the folded reading, the flat drawing is recognised on 187 of 200 sheets but the folded one on only 4, so only 2\% of sheets pass. Its flat CLIP score rises by 1.51 (0.88 to 2.17) and its folded score falls by 6.44 (5.43 to 7.44). Reading balance recognises the folded reading on 115 sheets instead of 104 and the flat reading on 147 instead of 144, so that 40.0\% of sheets pass instead of 31.0\%. With the crease loss, 115 folded readings are recognised instead of 106, and the pass rate is 40.0\% instead of 35.5\%. Ink stays between 5.7\% and 6.1\% of the sheet in these rows. None of the nine variants of \cref{tab:ablation,tab:design} recognises the folded reading on more sheets than \method.

\begin{table}[t]
\centering
\tablestyle{4pt}
\begin{tabular}{@{}lccccc@{}}
\toprule
Variant & Flat $\uparrow$ & V $\uparrow$ & Pass $\uparrow$ & Coverage $\uparrow$ & Ink $\downarrow$ \\
\midrule
w/o folded reading & \textbf{187} & 4 & 2.0 & 4.0 & 6.1 \\
w/o reading balance & 144 & 104 & 31.0 & 64.0 & \underline{5.8} \\
w/o crease loss & \underline{152} & \underline{106} & \underline{35.5} & \underline{68.0} & 6.1 \\
\midrule
\ours \method (full) & \ours 147 & \ours \textbf{115} & \ours \textbf{40.0} & \ours \textbf{72.0} & \ours \textbf{5.7} \\
\bottomrule
\end{tabular}
\caption{\textbf{Readings under each contribution} on the rows of \cref{tab:ablation}. Flat and V count the 200 sheets on which CLIP ranks that reading's prompt first among the 98 prompts. Pass rate, coverage and ink are in \%. Bold marks the best and underline the second.}
\label{tab:abl_read}
\end{table}

\noindent\textbf{Crease Loss at Other Thresholds:} \cref{sec:ablation} measures the ink that the two layers of a fold share, counting pixels below 0.5 in luminance. At thresholds of 0.25, 0.75 and 0.95, the layers share 31.0\%, 32.8\% and 32.8\% of the smaller layer's ink without the crease loss, against 22.9\%, 24.4\% and 24.4\% with it.

\section{More Qualitative Results}
\label{sec:new}

\noindent\textbf{New Pairs:} \cref{fig:newpairs} applies \method beyond the 50 pairs of the main paper, to objects that fold along either crease, to proper nouns and to scenes. None of their concepts appears in the 50 pairs. The horizontal fold lays the upper half over the lower one (\cref{sec:formulation}), so its folded concepts are wide and low, such as a helicopter or a tortoise. These sheets use the recipe of \cref{tab:recipe} with 128 strokes and 5,000 steps, each trained on the fold it shows.

\noindent\textbf{Qualitative Results:} When folded, the two halves of each flat drawing meet in the second drawing. The bands of a beehive become the tiers of a pagoda, the helmet of Darth Vader folds into R2-D2, Charlie Chaplin folds into the Oscar statuette and a mountain range folds into a city skyline.

\section{Physical Prototype}
\label{sec:proto}

\noindent\textbf{From Screen to Paper:} The fold of \cref{eq:fold} assumes that every layer stays visible, which a transparent sheet provides, so a FoldSketch drawing can be folded by hand. We redraw the butterfly and the face of \cref{fig:teaser} with a permanent marker on transparent plastic film, one drawing per sheet, and fold each sheet along its vertical centre line, laying the left half over the right one (\cref{fig:proto}). Flat, the two sheets show a butterfly and a face. Folded, they show an hourglass and a giraffe, the folded readings of \cref{fig:teaser}.

\section{Limitations and Future Work}
\label{sec:limit}

\noindent\textbf{Translucent Paper:} The fold of \cref{eq:fold} shows every layer, as on tracing paper, on a sheet held against the light or on the film of \cref{sec:proto}. On opaque paper the folded half hides the strokes beneath it, and the fold-in of MAD magazine~\cite{jaffee2011madfoldin} relies on this, joining the outer strips of the page. Modelling occlusion, where a flap hides what lies under it, would extend \method to ordinary paper and to Jaffee's fold-ins.

\noindent\textbf{Runtime:} Like other score-distillation methods, \method optimises each sheet, which takes 9.1 minutes on one GPU. A feed-forward model trained on such sheets could make fold drawings interactive.

\noindent\textbf{More Readings and Other Folds:} \cref{fig:three} shows sheets with three readings, and the fold operator of \cref{eq:operator} composes any sequence of creases. Longer sequences, creases off the centre and diagonal creases are natural next steps, as are the other many-to-one maps of the conclusion, rolling the paper and stacking cut pieces.

\end{document}